\documentclass[10pt,twocolumn,letterpaper]{article}

\usepackage[pagenumbers]{cvpr}      % To produce the REVIEW version
\definecolor{cvprblue}{rgb}{0.21,0.49,0.74}
\usepackage[pagebackref,breaklinks,colorlinks,allcolors=cvprblue]{hyperref}

\usepackage[table,xcdraw,dvipsnames]{xcolor}
\usepackage{wrapfig}
\usepackage{subcaption}
\usepackage{graphicx}
\usepackage{multirow} 
\usepackage{floatflt}
\usepackage{algorithm}
\usepackage{algorithmic}
\usepackage{bm} 
\usepackage[most]{tcolorbox} 
\usepackage{arydshln}
\usepackage{adjustbox}
\usepackage{makecell}
\usepackage{tablefootnote}
\def\paperID{XXXX} % *** Enter the Paper ID here
\def\confName{CVPR}
\def\confYear{2026}

\newcommand{\ours}{{PROGRESS }}
\title{Learning What Matters: \textcolor{orange}{Pr}ioritized C\textcolor{orange}{o}ncept Learnin\textcolor{orange}{g} via \textcolor{orange}{R}elative \textcolor{orange}{E}rror-driven \textcolor{orange}{S}ample \textcolor{orange}{S}election}

\author{
   Shivam Chandhok\thanks{Equal Contribution, order determined by coin flip. $\dagger$ Equal Advising.}\  \ $^{1,2,4}$, Qian Yang$^{\ast1,3}$, Oscar Mañas$^{1,3}$, Kanishk Jain$^{1,3}$, \\ {Leonid Sigal}$^{\dagger2,4,5}$, {Aishwarya Agrawal}$^{\dagger1,3,5}$
       \vspace{2mm}
    \\
  $^1$ Mila - Québec AI Institute
  $^2$ University of British Columbia \\ 
     $^3$ Université de Montréal $^4$ Vector Institute for AI 
   $^5$ Canada CIFAR AI Chair 
   \\
    \normalsize {\tt\small \{qian.yang,  aishwarya.agrawal\}@mila.quebec}~~~~ {\tt\small \{chshivam, lsigal\}@cs.ubc.ca}
}
\definecolor{lightgray}{gray}{0.9}
\definecolor{lightorange}{RGB}{255, 204, 153} 

\definecolor{lightgreen}{RGB}{32, 144, 140}  % Light Green
\definecolor{yellow}{RGB}{68, 190, 112} % Yellow
\definecolor{yellowgreen}{RGB}{189, 222, 38} % Yellow
\definecolor{darkblue}{RGB}{52,94,141} % Yellow

\definecolor{c1}{RGB}{180, 242, 230}  % Light Green
\definecolor{c2}{RGB}{137, 203, 225} % Yellow
\definecolor{c3}{RGB}{127, 168, 216} % Yellow
\definecolor{c4}{RGB}{120,135,204} % Yellow

\begin{document}
\maketitle

\begin{abstract}
Instruction tuning has been central to the success of recent vision-language models (VLMs), but it remains expensive—requiring large scale datasets, high-quality annotations and large-compute budget. We propose 
\textbf{PR}ioritized c\textbf{O}ncept learnin\textbf{G} via \textbf{R}elative \textbf{E}rror-driven \textbf{S}ample
\textbf{S}election -- \textcolor{orange}{\textbf{PROGRESS}} --
a data- and compute-efficient framework that enables VLMs to dynamically select what to learn next based on their evolving needs during training. At each stage, the model tracks its learning progress across skills and selects the most \emph{informative} samples: those it \emph{has not already mastered} and \emph{are not too difficult to learn} at the current state of training.
This strategy effectively controls skill acquisition and the order in which skills are learned. 
Specifically, we sample from skills showing the highest learning progress, prioritizing those with the most rapid improvement. 
Unlike prior methods, \ours requires no upfront answer annotations, querying answers only on a \emph{need basis}, avoids reliance on additional supervision from auxiliary VLM, or compute-heavy gradient computations for data selection. Experiments across multiple instruction-tuning datasets of varying scales demonstrate that \ours consistently outperforms state-of-the-art baselines with much less data and supervision. Additionally, we show strong cross-architecture generalization to different VLMs and transferability to larger models, validating \ours as a scalable solution for efficient learning.\footnote{Project website: \url{https://mylittlechange.github.io/PROGRESS_web/}}

  \end{abstract}
      
\section{Introduction}
\label{sec:intro}

Multimodal vision-language models (VLMs) such as GPT-4V~\citep{openai2024gpt4technicalreport}, Gemini~\citep{team2023gemini}, LLaVA~\citep{llava,llava-1.5}, and InternVL~\citep{chen2024internvl} demonstrate impressive general-purpose capabilities across tasks like image comprehension and visual question answering. Much of this success stems from large-scale fine-tuning on high-quality image-text corpora, particularly visual instruction-tuning (IT) datasets~\citep{zhang2023llavar,vision-flan}, which significantly enhance instruction-following and reasoning abilities. A growing trend in 
building stronger VLMs has been to simply scale up: collecting larger more diverse IT datasets with better annotations and using them to instruction-tune increasingly powerful models~\citep{chen2023sharegpt4v,llava-1.5}.

However, such pipelines are increasingly resource-intensive—annotation-heavy when relying on human-labeled supervision ({\it e.g.}, bounding boxes, object tags) and monetarily costly when generating instructions via proprietary models like GPT-4~\citep{llava,llava-1.5}, alongside significant computational overhead. 
These factors make such pipelines increasingly inaccessible to individual researchers and smaller academic labs. More importantly, it is even unclear whether the entirety of these large corpora is necessary for strong VLM performance. We posit that many samples are redundant or uninformative, and that comparable results could be achieved using fewer, informative samples.

To this end, we investigate how to select the \emph{most informative} visual instruction-tuning (IT) samples based on the model’s own evolving learning state. We ask: \textit{Can VLMs indicate what they can most effectively learn at a give stage of training?} Inspired by curriculum learning, we develop a framework in which the model periodically self-evaluates its current knowledge and identifies the skills it is ready to acquire next—those that would most benefit its learning progress. Specifically, we track the relative change in skill performance across iterations to estimate where learning improves fastest, encouraging the model to prioritize these skills. We hypothesize that this enables the VLM to actively select training samples that are most informative: those that are \emph{not already mastered} by the model, and are \emph{not too difficult}, allocating its limited budget to those offering the \emph{highest learning benefit and thereby improving overall training efficiency}. Overall, \ours adapts to the model’s evolving learning state and helps acquire essential skills while \emph{balancing  informativeness of selected samples with diversity of concepts}—a property crucial for covering important data modes and supporting generalization.

{Experimental results across multiple instruction-tuning datasets of varying scale demonstrate that \ours achieves up to \textbf{99–100\%} of full-data performance using only \textbf{16–20\%} of the labeled training data, \textbf{while also being faster in overall training time}—including all self-evaluation overhead}.
In addition to these gains, \ours offers several practical advantages over existing approaches. First, unlike static scoring-based methods \citep{el2n, entropy, perplexity, clipscore} or concept-driven strategies that rely on additional reference VLMs~\citep{coincide}, our method uses dynamic feedback from the model’s own learning progress to guide training. Second, while many prior methods assume full access to ground-truth annotations upfront~\citep{coincide, icons}, we operate in a more realistic setting where the training pool is initially unlabeled and answers are queried only on a \textit{need basis}—drastically reducing annotation cost. Third, instead of merely selecting \textit{which} samples to train on~\citep{coincide, icons}, our approach also decides \textit{when} to introduce each skill—enabling curriculum-style control over both skill acquisition and learning order.
Our contributions are as follows:

\begin{itemize}
    \item \textbf{We propose PROGRESS}, a dynamic, progress-driven framework that selects the most informative samples for VLM instruction tuning by leveraging the model’s \textbf{own learning progress} across automatically discovered multimodal concepts—\textbf{without relying on auxiliary VLMs}, manual heuristics, or full-data supervision.

    \item \ours is 
    \textbf{data and label efficient} --  
    achieves near full-data performance (\textbf{99--100\%}) using only \textbf{16--20\%} labeled data, generalizing across datasets, architectures, and model scales (LLaVA-v1.5-7B/13B, Qwen2-VL). 
    
    \item \ours is also \textbf{faster in total training time}, including the self-evaluation overhead.

    \item We analyze \emph{what} the model learns and \emph{when}, revealing an interesting  curriculum over concept types and difficulty, offering new insights into efficient VLM training.
\end{itemize}

% \vspace{-0.1in}
\section{Related Work}
\label{related_work}

\begin{figure}
    \centering
    \includegraphics[width=\linewidth]{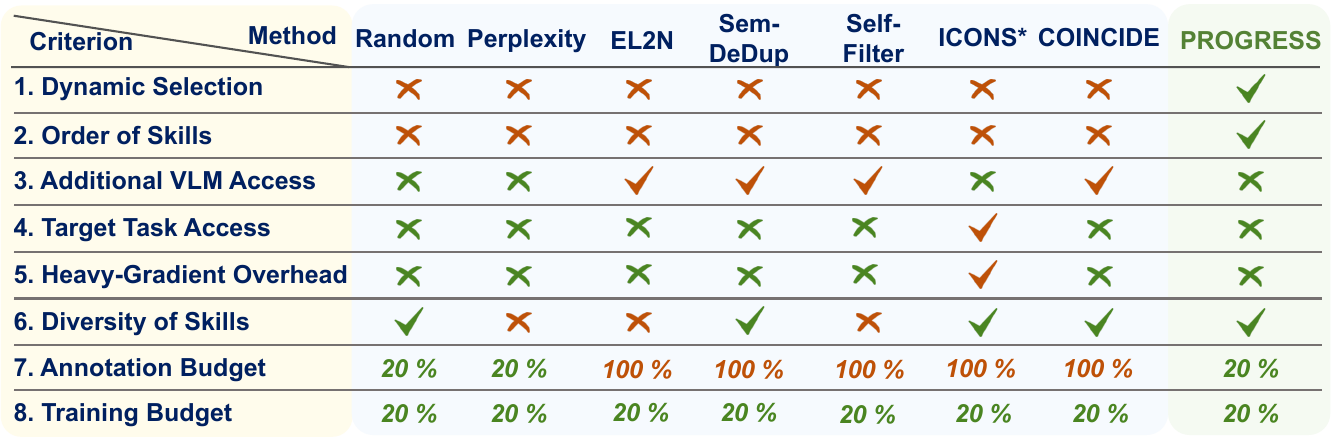}
    \caption{\textbf{Comparison with Prior Efficient Learning Methods for VLMs.} \textcolor{OliveGreen}{\textbf{Green}} denote \textcolor{OliveGreen}{\textbf{desirable}} properties for efficient learning, while \textcolor{BrickRed}{\textbf{Red}} indicate \textcolor{BrickRed}{\textbf{limitations}}. PROGRESS \textit{satisfies all key desirable criteria} while requiring only 20\% data. See {Appendix \ref{app:baselines}} for details of prior approaches.
    % we compare with.
    }
    \label{fig:comparison}
    % \vspace{-5mm}
\end{figure}

% \vspace{-3pt}
\textbf{Data Efficient Learning for VLMs.}
Prior approaches to efficient VLM training (see Fig.~\ref{fig:comparison}) typically select a coreset—{\em i.e.}, a representative subset of the data for training—using static metrics such as CLIP-Score~\citep{clipscore}, EL2N~\citep{el2n}, perplexity~\citep{perplexity}, entropy~\citep{entropy}, or by training auxiliary scoring networks~\citep{Chen2024selffilter}. 
However, these methods perform one-time sample selection before training and cannot adapt to the model’s evolving needs. Static score metrics also miss important data modes, leading to poor diversity and reduced generalization~\citep{coincide,adapt-inf}, and in some cases even underperform random selection~\citep{coincide}.
Gradient-based methods~\citep{icons,tive}, while more principled, are computationally prohibitive—requiring large memory to store high-dimensional gradients and hundreds of GPU hours (e.g., ICONS\footnote{Unpublished concurrent work; we reproduce and compare in {Appendix~\ref{app:icons}}}~\citep{icons})—contradicting the very goal of efficient training. Some also assume access to explicit knowledge of the target task or distribution through labeled samples, as in ICONS~\citep{icons}, which is rarely realistic in general-purpose VLM settings. {More recently, prominent work has explored selection using additional reference VLMs—auxiliary models that themselves require large-scale instruction tuning. 
A notable example is COINCIDE~\citep{coincide}, which extracts internal activations from a separately trained VLM (e.g., TinyLLaVA~\citep{zhou2024tinyllava}) to guide coreset selection. 
% However, our framework has several key differences over COINCIDE as explained below.
However, COINCIDE exhibits \textbf{\emph{several limitations}}: it requires a fully trained additional auxiliary VLM, performs static one-time selection without controlling the order of skills being learned, needs ground-truth annotations for the entire dataset, and requires manual human inspection to first select appropriate activations—all of which make it resource-intensive and difficult to scale.}

% % \vspace{-0.05in}

\begin{figure*}[t]
    \centering
    \includegraphics[width=0.9\textwidth]{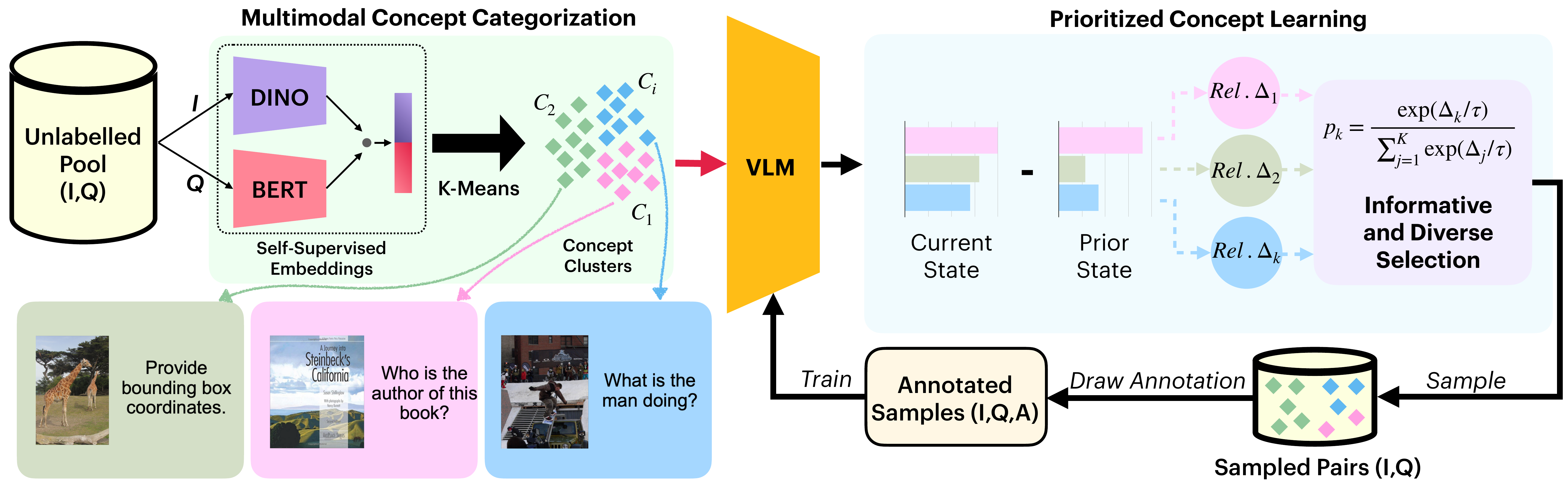}
\caption{\textbf{Overall Pipeline.} Our framework consists of two stages: (1) \textit{Multimodal Concept Categorization}, which partitions the unlabeled pool \( \mathbb{U} \) into distinct skills by assigning each sample \( (I, Q) \) to a specific skill cluster, and (2) \textit{Prioritized Concept Learning}, where the model actively selects the most informative samples—those showing the highest improvement in its objective (e.g., accuracy or loss) relative to its prior state. We query annotations for only these selected samples on a \textit{need basis}, forming labeled set \( (I, Q, A) \), which is used for training.
}
\label{fig:trainingpipe}
% \vspace{-0.15in}
\end{figure*}

\noindent
\textbf{Curriculum Learning, Self-Paced Learning, and Active Learning.}
{Curriculum learning improves generalization by ordering data from easy to hard~\citep{bengio2009curriculum}, while self-paced learning adapts this order based on the learner’s progress~\citep{kumar2010self,sachan2016easy}. These ideas have been explored in NLP~\citep{sachan2016easy,mindermann2022prioritized} and in controlled synthetic multimodal settings~\citep{lba}, but typically with small-scale models and externally designed heuristics. To the best of our knowledge, \ours is the \emph{\textbf{first method}} to show how these principles can be used to train large-scale VLMs efficiently. \ours shares the overall goal of active learning in selecting informative samples, but rather than relying on instance-level uncertainty, it prioritizes samples where the model shows the greatest relative improvement—yielding a progress-driven, curriculum-like ordering instead.}
% of a static uncertainty-based selection.} 

\noindent
\textbf{{Key Differences Over Prior Methods.}} 
{\ours introduces a novel framework that uses the \textbf{model’s own learning progress} to select the most informative samples for data-efficient training of large VLMs, \textbf{in contrast} to COINCIDE, which relies on a separately trained auxiliary VLM. \ours \textbf{dynamically adapts} to the model’s evolving learning state, explicitly controlling both which skills are acquired and the order in which they are learned (\textit{rows 1–2} in Fig.~\ref{fig:comparison}), \textbf{whereas} COINCIDE performs a static, one-shot selection that ignores acquisition order. In \ours supervision is applied strictly on a \textit{need-to-label} basis, requiring annotations for only $\sim$20\% of the dataset (\textit{row 7}), \textbf{while} COINCIDE assumes full-data labels. Moreover, \ours requires no target-task knowledge (\textit{row 4}), avoids compute-heavy gradient estimation (\textit{row 5}), and maintains diverse skill coverage (\textit{row 6}). \textbf{To the best of our knowledge}, \ours is the \textbf{first} method to use a model’s own signal to drive a \textbf{self-paced, progress-driven curriculum} for efficient large-VLM training—focusing learning where progress is greatest while jointly controlling both skill acquisition and learning order.}
% \vspace{-0.1in}

\section{Problem Setting and Overall Framework}
\label{sec:method}
% \vspace{-0.05in}

% % \vspace{-0.10in}
\textbf{Problem Setting.}
We now formally introduce the data-efficient learning  setting for training VLMs. We denote an image by \( I \), a question by \( Q \), forming an image-question pair \( (I, Q) \in \mathbb{U} \), where \( \mathbb{U} \) is an unlabeled pool of such pairs. Unlike previous efficient learning methods, we do not assume access to the corresponding answers \( A \in \mathbb{A} \) for all pairs in \( \mathbb{U} \), and thus refer to this pool as unlabeled. The learner is provided with: (1) the unlabeled pool \( \mathbb{U} \); and (2) a fixed answer budget \( b \), specifying the maximum number of pairs from \( \mathbb{U} \) for which it can query an answer \( A \in \mathbb{A} \) and use for training, where \( |\mathbb{A}| = b \ll |\mathbb{U}| \). The goal is to learn a vision-language model \( \mathtt{VLM}(A \mid I, Q) \) that can accurately predict an answer for a new image-question pair, while only using \( b \) selected and labeled samples during training. The central challenge lies in identifying the \emph{most informative} \( (I, Q) \) pairs to annotate within the constrained budget \( b \), such that the resulting model trained on these \( (I, Q, A) \) pairs performs comparably to one trained on the fully labeled dataset. This setup \textbf{\textit{follows standard setting}} used in prior data-efficient learning~\citep{coincide,icons,Chen2024selffilter}

% % \vspace{-2pt}
\textbf{Overall Framework.} Our overall framework % for efficient training of VLMs 
is shown in Figure \ref{fig:trainingpipe}. We employ a two-stage pipeline:
% % \vspace{-0.05in}

% % \vspace{-5pt}
\begin{enumerate}
    \item[(1)] \underline{Multimodal Concept Categorization.} Given an unlabeled data pool \( \mathbb{U} \) with image-question pairs \( (I, Q) \in \mathbb{U} \), we first partition \( \mathbb{U} \) into \( K \) concept clusters in a \emph{fully unsupervised} manner, assigning each sample \( (I, Q) \) to a specific concept. This enables tracking the model's progress on individual concepts and supports a self-paced training strategy where the model's own learning signals determines which concepts to prioritize next.

     \item[(2)]\underline{Prioritized Concept Learning.} During training, the model periodically self-evaluates its knowledge by comparing its current performance to prior state, identifying skills where performance improves fastest relative to prior state. Samples \( (I, Q) \) from these skills are then selected and answer annotations \( A \in \mathbb{A} \) are queried only for these selected samples.
\end{enumerate}

% % \vspace{-3mm}
{To obtain reliable skill-level performance estimates at the start of training—when the model is still untrained—we include a brief \textit{warmup phase}, consistent with prior work.
We adopt simple sampler, selecting samples from our unsupervised concept clusters with probability:
$
P_i \propto \exp\left({S_i}/{\tau D_i}\right)$
where $S_i$ denotes the cluster's \textit{transferability} and $D_i$ its \textit{density} \cite{coincide} (see details in Appendix~\ref{sec:warmup}).
The warmup samples, together with those selected by our prioritized strategy, constitute the total annotation budget $b$, ensuring that the labeled set $(I,Q,A)$ \textbf{never exceeds this budget}. Training with the warmup set alone yields substantially worse performance, highlighting the importance of our core contribution -- progress-driven sampling (Table~\ref{tab:ablation_warmup}).}

% \vspace{-0.05in}
\subsection{Multimodal Concept Categorization}
\label{sec:MM_cat}
\begin{figure}
    \centering
    \includegraphics[width=\linewidth]{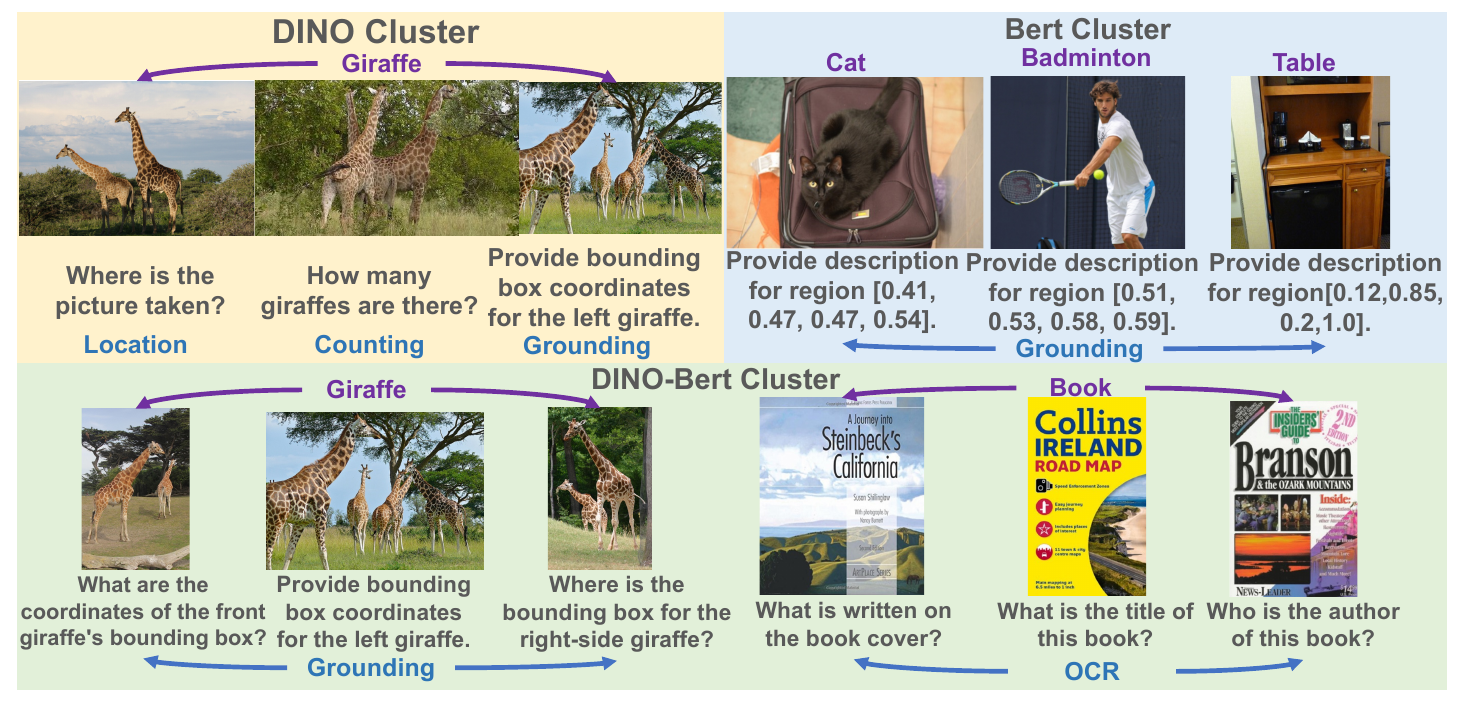}
    \caption{{\bf Cluster Visualization.} 
    Clustering with multimodal DINO-BERT features ensures purer skill clusters with higher intra-cluster
    and lower inter-cluster similarity compared to uni-modal partitioning. See \textbf{Word Cloud Visualization} {Appendix \ref{app:word_cloud}}. }
    \label{fig:cluster}
% \vspace{-0.2in}
\end{figure}

{We first get a set of \textbf{concept clusters} from the unlabeled instruction-tuning pool $U$. Instead of manually defining categories, we aim to obtain \textbf{coherent visual--linguistic groupings} where samples share similar image content and question type; we refer to these groupings as \textbf{concepts} or \textbf{skills}. For each image--question pair $(I,Q)$, we extract \textbf{self-supervised} features using a frozen DINO vision encoder $f_{\text{img}}(I)$ and a BERT question encoder $f_{\text{text}}(Q)$, concatenate and L2-normalize them:
\(
z(I,Q)=\operatorname{norm}\big([f_{\text{img}}(I);\; f_{\text{text}}(Q)]\big).
\)
We then apply \textbf{spherical k-means} to partition the pool into $K$ concept clusters $\{C_1,\dots,C_K\}$, each serving as a \textbf{concept group} for tracking learning progress. Using both modalities yields \textbf{more coherent and interpretable clusters} than unimodal variants; as shown in Fig.\ref{fig:cluster} and Appendix \ref{app:word_cloud} (Word Cloud Visualization), these groups often align with abilities such as object grounding, OCR, programming and multilingual tasks. Unlike COINCIDE---our closest prior competitor---which requires activations from an additional VLM, full annotations, and manual inspection, our categorization is \textbf{fully unsupervised \& model-agnostic}, requiring no annotations, auxiliary models, or human intervention.}

% \vspace{-0.05mm}
\subsection{Prioritized Concept Learning: Can VLMs indicate what they can most effectively learn at a give stage of training?}
\label{PCL}
% \vspace{-1.5pt}
{
Our goal is to guide the VLM toward the skills it can most effectively learn at its current stage of training. To use the annotation budget efficiently, we aim to select samples that offer high learning benefit—avoiding those that are too easy (redundant, low gain) or too hard (beyond the model’s current capacity). Concretely, we leverage the model’s \textbf{own learning progress} to identify informative samples: those that yield the \textbf{largest improvement} in the model’s objective (e.g., \textbf{accuracy} or \textbf{loss}) relative to its prior state.}

Formally, given an unlabeled pool $\mathbb{U} = \{(I,Q)\}$ partitioned into concept clusters $\mathcal{C}=\{C_1,\dots,C_K\}$, we track the model’s learning state at step $t$ using its accuracy $\text{Acc}_k^{(t)}$ on each cluster $k$, computed over the labeled training data observed so far. The \textbf{\textit{relative change in accuracy}} quantifies learning progress and drives sample selection. For each cluster, we compute improvement between steps $t$ and $t-\gamma$:
\begin{equation}
\Delta_k = \frac{\text{Acc}_k^{(t)} - \text{Acc}_k^{(t-\gamma)}}{\text{Acc}_k^{(t-\gamma)} + \epsilon},
\label{eq:delta}
\end{equation}
where $\epsilon$ ensures stability. The score $\Delta_k$ measures how rapidly the model is improving on cluster $k$ and serves as a proxy for \textit{sample informativeness}. Prioritizing high-$\Delta_k$ clusters yields a \textit{self-paced curriculum} that adapts to the model’s evolving capabilities, focusing learning where \textit{\textbf{progress is greatest}} - controlling both the acquisition of skills and the order in which they are learned. For exploration, we additionally sample $\delta\%$ of points uniformly at random, following prior curriculum-learning work~\citep{kumar2010self,lba}.

Annotations are queried only for the selected samples, forming the labeled set $(I,Q,A)$. This \textit{need-based annotation} strategy removes the requirement for full-dataset supervision used in prior coreset methods~\citep{coincide}, enabling far more scalable and efficient training.

However, selecting samples only from the highest-improvement cluster  hurt diversity by concentrating on a narrow skill set and leads to mode collapse—an issue also noted in prior work~\citep{coincide}. To mitigate this, we propose to sample from multiple high-$\Delta_k$ clusters using a \textit{temperature-controlled softmax}:
\begin{equation}
p_k = \frac{\exp(\Delta_k / \tau)}{\sum_{j=1}^K \exp(\Delta_j / \tau)} .
\label{eq:softmax}
\end{equation}

\noindent
Here, $p_k$ is the probability of sampling cluster $k$, and $\tau$ controls the sharpness of the distribution. Lower $\tau$ concentrates sampling on top-improving clusters (high informativeness but low diversity), while higher $\tau$ broadens sampling \& improves skill coverage. Balancing \textbf{informativeness vs.\ diversity} is essential for learning ( Fig.~\ref{fig:abl_temp}). At each step $t$, the sampling budget is allocated proportionally to $p_k$, \& only selected samples are annotated as $(I,Q,A)$ for training.

% \vspace{-4mm}

\begin{table*}[th]
    \tiny
    \caption{Comparison of coreset selection techniques for training LLaVA-v1.5 on the LLaVA-665K dataset using 20\% sampling ratio.  {\textbf{\textit{AL Only}} refers to methods which align with active learning principles.    \textbf{\textit{AL \& CL}} refers to methods which use combination of active and curriculum learning principles.} \colorbox{lightorange!40}{Methods highlighted in orange} require \textbf{additional} reference VLMs and 100\% dataset annotations for coreset selection, while {\colorbox{green!10}{methods highlighted in light green}} do not require either and thus considered better. The benchmark results are highlighted with \colorbox{c4!50}{best} and \colorbox{c4!15}{second best} within the respective categories (i.e, with and without utilizing additional information). The best and the second best relative score are in \textbf{bold} and \underline{underlined}, respectively. Reported numbers are averaged over 3 runs.} 
    \centering
    \resizebox{\textwidth}{!}
    {
        \renewcommand{\arraystretch}{1.2}
        \renewcommand{\tabcolsep}{3.0pt}
         \adjustbox{max width=0.9\textwidth}{
        \begin{tabular}{l c c c c c c c cc c c c c c c}
             \toprule
              {\textbf{Method}} & {\textbf{VQAv2}} & {\textbf{GQA}} & {\textbf{VizWiz}} & {\textbf{SQA-I}} & {\textbf{TextVQA}} & {\textbf{POPE}} & {\textbf{MME}} & \multicolumn{2}{c}{\textbf{MMBench}} & {\textbf{LLaVA-}} & {\textbf{SEED}} & {\textbf{AI2D}}   & {\textbf{ChartQA}} & {\textbf{CMMMU}}  &  \cellcolor{gray!25}{\textbf{Rel. (\%)}}\\
             & & & & & & & & {\textbf{en}} & {\textbf{cn}} & {\textbf{Wild}} & & & & \\
             \midrule
             \multicolumn{15}{c}{\textbf{LLaVA-v1.5-7B}} \\
             \midrule
             \ 0\ \ \ Full-Finetune &
             {\scriptsize 79.1} & {\scriptsize 63.0} & {\scriptsize 47.8} & {\scriptsize 68.4} & {\scriptsize 58.2} & {\scriptsize 86.4} & {\scriptsize 1476.9} & {\scriptsize 66.1} & {\scriptsize 58.9} & {\scriptsize 67.9}  & {\scriptsize 67.0} & {\scriptsize 56.4} & {\scriptsize 16.4} & {\scriptsize 22.1} & \cellcolor{gray!25}{\scriptsize 100}\\
             \cmidrule{0-15}
             
             \cellcolor{lightorange!40}  1\ \ \ Self-Sup (\textbf{\textit{AL Only}}) &
             {\scriptsize 74.9} & {\scriptsize \cellcolor{c4!15}59.5} & {\scriptsize 46.0} & {\scriptsize 67.8} & {\scriptsize 49.3} & {\scriptsize 83.5} & {\scriptsize 1335.9} & {\scriptsize 61.4} & {\scriptsize 53.8} & {\scriptsize 63.3} & {\scriptsize 62.5} & {\scriptsize 52.9} & {\scriptsize \cellcolor{c4!15}16.1} & {\scriptsize 23.4} & \cellcolor{gray!25}{\scriptsize 94.6}\\
             \cellcolor{lightorange!40}    2\ \ \ Self-Filter (\textbf{\textit{AL \& CL}}) &
             {\scriptsize 73.7} & {\scriptsize 58.3} & \cellcolor{c4!50}{\scriptsize 53.2} & {\scriptsize 61.4} & {\scriptsize 52.9} & {\scriptsize 83.8} & {\scriptsize 1306.2} & {\scriptsize 48.8} & {\scriptsize 45.3} & {\scriptsize 64.9} & {\scriptsize 60.5} & {\scriptsize 48.7} & {\scriptsize 14.1} & {\scriptsize 19.8} & \cellcolor{gray!25}{\scriptsize 90.1}\\
             \cellcolor{lightorange!40}  3\ \ \ EL2N (\textbf{\textit{AL Only}}) &
             {\scriptsize \cellcolor{c4!15} 76.2} & {\scriptsize 58.7} & {\scriptsize 43.7} & {\scriptsize 65.5} & {\scriptsize 53.0} & {\scriptsize 84.3} & {\scriptsize \cellcolor{c4!15}1439.5} & {\scriptsize 53.2} & {\scriptsize 47.4} & {\scriptsize 64.9} & {\scriptsize   61.8}& {\scriptsize 49.3} & {\scriptsize \cellcolor{c4!50}16.5}  & {\scriptsize 23.9} & \cellcolor{gray!25}{\scriptsize 93.4 }\\
             \cellcolor{lightorange!40} 4\ \ \ SemDeDup &
             {\scriptsize 74.2} & {\scriptsize 54.5} & {\scriptsize \cellcolor{c4!15}46.9} & {\scriptsize 65.8} & {\scriptsize \cellcolor{c4!15}55.5} & {\scriptsize 84.7} & {\scriptsize 1376.9} & {\scriptsize 52.2} & {\scriptsize 48.5} & {\scriptsize  \cellcolor{c4!50}70.0} & {\scriptsize 60.9} & {\scriptsize  \cellcolor{c4!50}53.5} & {\scriptsize  15.8}  & {\scriptsize \cellcolor{c4!15}24.2} &\cellcolor{gray!25}{\scriptsize 94.1}\\
             \cellcolor{lightorange!40}    5\ \ \ D2-Pruning &
             {\scriptsize 73.0} & {\scriptsize 58.4} & {\scriptsize 41.9} & {\scriptsize \cellcolor{c4!50}69.3} & {\scriptsize 51.8} & {\scriptsize \cellcolor{c4!15}85.7} & {\scriptsize 1391.2} & {\scriptsize  \cellcolor{c4!50}65.7} & {\scriptsize  \cellcolor{c4!50}57.6} & {\scriptsize 63.9} & {\scriptsize \cellcolor{c4!15}62.1} & {\scriptsize 52.5} & {\scriptsize 15.3} & {\scriptsize 22.3} & \cellcolor{gray!25}{\scriptsize 94.8 }\\
             \cellcolor{lightorange!40}    6\ \ \ COINCIDE (\textbf{\textit{AL Only}})  &{\scriptsize \cellcolor{c4!50}76.5} & {\scriptsize \cellcolor{c4!50}59.8} & {\scriptsize 46.8} & {\scriptsize \cellcolor{c4!15}69.2} & {\scriptsize \cellcolor{c4!50}55.6} & {\scriptsize \cellcolor{c4!50}86.1} & {\scriptsize \cellcolor{c4!50}1495.6} & {\scriptsize \cellcolor{c4!15} 63.1} & {\scriptsize \cellcolor{c4!15}54.5} & {\scriptsize \cellcolor{c4!15}67.3}  & {\scriptsize \cellcolor{c4!50}62.3} & {\scriptsize \cellcolor{c4!15}53.3} & {\scriptsize \cellcolor{c4!15}16.1} & {\scriptsize  \cellcolor{c4!50}24.3} & \cellcolor{gray!25}{\scriptsize 97.8} \\
         \cmidrule{0-15}
            \cellcolor{green!10} 7\ \ \ Random &   {\scriptsize \cellcolor{c4!15}75.7} & {\scriptsize  \cellcolor{c4!50}58.9} & {\scriptsize 44.3} & {\scriptsize 68.5} & {\scriptsize  \cellcolor{c4!50}55.3} & {\scriptsize 84.7} & {\scriptsize 1483.0} & {\scriptsize  \cellcolor{c4!15}62.2} &  \cellcolor{c4!15}{\scriptsize 54.8} & {\scriptsize 65.0}& {\scriptsize 61.7} & {\scriptsize 50.2} & {\scriptsize \cellcolor{c4!15}15.1}  & {\scriptsize 21.9} & \cellcolor{gray!25}{\scriptsize 95.0}\\
             \cellcolor{green!10}  8\ \ \ CLIP-Score &
             {\scriptsize 73.4} & {\scriptsize 51.4} & {\scriptsize 43.0} & {\scriptsize 65.0} & {\scriptsize 54.7} & {\scriptsize 85.3} & {\scriptsize 1331.6} & {\scriptsize 55.2} & {\scriptsize 52.0} & {\scriptsize \cellcolor{c4!15}66.2} & {\scriptsize 61.0} & {\scriptsize 49.1} & {\scriptsize 14.3} & {\scriptsize 20.3} & \cellcolor{gray!25}{\scriptsize 90.6}\\
              \cellcolor{green!10}  9\ \ \ Perplexity (\textbf{\textit{AL Only}}) &
             {\scriptsize  \cellcolor{c4!50}75.8} & {\scriptsize 57.0} & {\scriptsize 47.8} & {\scriptsize 65.1} & {\scriptsize 52.8} & {\scriptsize 82.6} & {\scriptsize 1341.4} & {\scriptsize 52.0} & {\scriptsize 45.8} & {\scriptsize  \cellcolor{c4!50}68.3} & {\scriptsize 60.8} & {\scriptsize 48.7} & {\scriptsize 14.5} & {\scriptsize 20.9} & \cellcolor{gray!25}{\scriptsize 91.1}\\
            \cellcolor{green!10} \textbf{\ours (\textit{AL \& CL}) } \\ 
             \cellcolor{green!10}  10\ \ \ Loss as Obj. & {\scriptsize \cellcolor{c4!15}75.7} & {\scriptsize 58.6} & {\scriptsize \cellcolor{c4!15}49.6} & {\scriptsize \cellcolor{c4!50}70.1} & {\scriptsize \cellcolor{c4!15}55.1} & {\scriptsize \cellcolor{c4!50}86.3} & {\scriptsize \cellcolor{c4!50}1498.4} & {\scriptsize \cellcolor{c4!50}62.5} & {\scriptsize \cellcolor{c4!50}55.5} & {\scriptsize 65.5} & {\scriptsize \cellcolor{c4!50}63.4} & {\scriptsize \cellcolor{c4!50}53.3} & {\scriptsize \cellcolor{c4!50}17.3} & {\scriptsize \cellcolor{c4!15}23.7} & \cellcolor{gray!25}{\scriptsize \underline{98.4}} \\
             \cellcolor{green!10}  11\ \ \  Accuracy as Obj. & {\scriptsize 75.2} & {\scriptsize \cellcolor{c4!15}58.8} & \cellcolor{c4!50}{\scriptsize 53.4} & \cellcolor{c4!15}{\scriptsize 69.9} & {\scriptsize \cellcolor{c4!15}55.1} & {\scriptsize \cellcolor{c4!15}85.9} & {\scriptsize \cellcolor{c4!15}1483.2} & {\scriptsize 61.1} & {\scriptsize 54.4} & {\scriptsize 65.5} & {\scriptsize \cellcolor{c4!15}63.0} & {\scriptsize \cellcolor{c4!15}52.8} & {\scriptsize  \cellcolor{c4!50}17.3}  & {\scriptsize  \cellcolor{c4!50}24.6} & \cellcolor{gray!25}{\scriptsize \textbf{98.8}} \\
                \midrule
                \multicolumn{15}{c}{\textbf{LLaVA-v1.5-13B}} \\
                \midrule

                \ 12\ \ \ Full-Finetune &
                {\scriptsize 80.0} & {\scriptsize 63.3} & {\scriptsize 58.9} & {\scriptsize 71.2} & {\scriptsize 60.2} & {\scriptsize 86.7} & {\scriptsize 1541.7} & {\scriptsize 68.5} & {\scriptsize 61.5} & {\scriptsize 69.5} &  {\scriptsize 68.3} &{\scriptsize 60.1}	 & {\scriptsize 19.3}	 &  {\scriptsize 22.1} & \cellcolor{gray!25}{\scriptsize 100}\\
                \cmidrule{0-15}
                  \cellcolor{lightorange!40}    13\ \ \ Self-Sup (\textbf{\textit{AL Only}}) &
                {\scriptsize 76.3} & {\scriptsize \cellcolor{c4!50}60.5} & {\scriptsize \cellcolor{c4!15}50.0} & {\scriptsize 70.2} & {\scriptsize 52.7} & {\scriptsize \cellcolor{c4!15}85.4} & {\scriptsize 1463.8} & {\scriptsize 63.7} & {\scriptsize\cellcolor{c4!15}57.6} & {\scriptsize 64.9} & {\scriptsize 65.2} & {\scriptsize 53.3} & {\scriptsize \cellcolor{c4!15}17.2} & {\scriptsize 23.2} & \cellcolor{gray!25}{\scriptsize 93.8}\\
               \cellcolor{lightorange!40}     14\ \ \ Self-Filter (\textbf{\textit{AL \& CL}}) &
                {\scriptsize 75.0} & {\scriptsize \cellcolor{c4!15}59.8} & {\scriptsize 48.6} & {\scriptsize 69.5} & {\scriptsize 55.8} & {\scriptsize 84.5} & {\scriptsize 1446.9} & {\scriptsize 58.8} & {\scriptsize 51.8} & {\scriptsize {\cellcolor{c4!50}69.1}} & {\scriptsize 65.3} & {\scriptsize 52.4} & {\scriptsize 16.9} & {\scriptsize 23.1} & \cellcolor{gray!25}{\scriptsize 92.6}\\
                 \cellcolor{lightorange!40} 15\ \ \ EL2N (\textbf{\textit{AL Only}}) &
                {\scriptsize {\cellcolor{c4!15}77.2}} & {\scriptsize 59.6} & {\scriptsize {\cellcolor{c4!50}54.8}} & {\scriptsize 69.9} & {\scriptsize 56.1} & {\scriptsize 84.1} & {\scriptsize {\cellcolor{c4!15}1531.0}} & {\scriptsize 59.3} & {\scriptsize 52.3} & {\scriptsize 65.8} & {\scriptsize \cellcolor{c4!15}65.7} & {\scriptsize \cellcolor{c4!50}53.9} & {\scriptsize 17.0} & {\scriptsize 24.4} & \cellcolor{gray!25}{\scriptsize 94.4}\\
                \cellcolor{lightorange!40}    16\ \ \ SemDeDup &
                {\scriptsize 75.6} & {\scriptsize 57.5} & {\scriptsize 48.3} & {\scriptsize {\cellcolor{c4!50}70.5}} & {\scriptsize {\cellcolor{c4!15}57.7}} & {\scriptsize 85.3} & {\scriptsize 1397.6} & {\scriptsize 59.0} & {\scriptsize 51.1} & {\scriptsize \cellcolor{c4!15}68.7} & {\scriptsize 64.9} & {\scriptsize 53.2} & {\scriptsize 16.8} & {\scriptsize \cellcolor{c4!15}24.6} & \cellcolor{gray!25}{\scriptsize 92.9}\\
                \cellcolor{lightorange!40}   17\ \ \ D2-Pruning &
                {\scriptsize 73.9} & {\scriptsize {\cellcolor{c4!50}60.5}} & {\scriptsize 49.8} & {\scriptsize {\cellcolor{c4!15}70.4}} & {\scriptsize 55.2} & {\scriptsize 84.9} & {\scriptsize 1463.0} & {\scriptsize {\cellcolor{c4!50}67.3}} & {\scriptsize \cellcolor{c4!50}{59.9}} & {\scriptsize 66.5} & {\scriptsize \cellcolor{c4!50}65.9} & {\scriptsize \cellcolor{c4!15}53.4} & {\scriptsize 16.9} & {\scriptsize 23.5} & \cellcolor{gray!25}{\scriptsize 94.7}\\
                 \cellcolor{lightorange!40}    18\ \ \ COINCIDE (\textbf{\textit{AL Only}}) & {\scriptsize \cellcolor{c4!50}77.3} & {\scriptsize 59.6} & {\scriptsize 49.6} & {\scriptsize 69.2} & {\scriptsize \cellcolor{c4!50}58.0} & {\scriptsize \cellcolor{c4!50}87.1} & {\scriptsize\cellcolor{c4!50} 1533.5} & {\scriptsize \cellcolor{c4!15}64.5} & {\scriptsize 56.6} & {\scriptsize 66.4}& {\scriptsize \cellcolor{c4!50}65.9}  & {\scriptsize 52.9}  & {\scriptsize \cellcolor{c4!50}18.4}    & {\scriptsize \cellcolor{c4!50} 25.0 } & \cellcolor{gray!25}{\scriptsize 95.9}\\
                    \cmidrule{0-15}
               \cellcolor{green!10}   19\ \ \ Random &
                {\scriptsize 76.7} & {\scriptsize \cellcolor{c4!50}{60.5}} & {\scriptsize 48.0} & {\scriptsize 68.8} & {\scriptsize {\cellcolor{c4!15}57.7}} & {\scriptsize 84.8} & {\scriptsize 1484.9} & {\scriptsize 62.8} & {\scriptsize 55.2} & {\scriptsize 68.6} & {\scriptsize \cellcolor{c4!50}65.5} & {\scriptsize\cellcolor{c4!15}57.9}   & {\scriptsize 17.1}   & {\scriptsize 24.3}  & \cellcolor{gray!25}{\scriptsize 95.0}\\
                \cellcolor{green!10}   20\ \ \ CLIP-Score &
                {\scriptsize 75.3} & {\scriptsize 52.6} & {\scriptsize 42.2} & {\scriptsize 69.7} & {\scriptsize 57.3} & {\scriptsize {85.4}} & {\scriptsize 1426.3} & {\scriptsize 60.4} & {\scriptsize 54.0} & {\scriptsize 68.1} & {\scriptsize 63.3} & {\scriptsize 52.8} & {\scriptsize 17.4} & {\scriptsize 23.7} & \cellcolor{gray!25}{\scriptsize 91.8}\\
                \cellcolor{green!10}  21\ \ \ Perplexity (\textbf{\textit{AL Only}}) &
                {\scriptsize \cellcolor{c4!50}{77.0}} & {\scriptsize 58.5} & {\scriptsize 48.2} & {\scriptsize 68.7} & {\scriptsize 54.8} & {\scriptsize 83.1} & {\scriptsize \cellcolor{c4!50}1508.8} & {\scriptsize 57.5} & {\scriptsize 50.3} & {\scriptsize \cellcolor{c4!15}68.7} & {\scriptsize 64.7} & {\scriptsize 53.1} & {\scriptsize 17.6} & {\scriptsize 23.8} & \cellcolor{gray!25}{\scriptsize 92.7}\\
                 \cellcolor{green!10}   \textbf{\ours (\textit{AL \& CL})}  \\ 
                 \cellcolor{green!10}    22\ \ \ Loss as Obj. & {\scriptsize 76.8} & {\scriptsize \cellcolor{c4!15}59.7} & {\scriptsize \cellcolor{c4!50}54.6} & {\scriptsize \cellcolor{c4!50}70.4} & {\scriptsize \cellcolor{c4!50}58.0} & {\scriptsize \cellcolor{c4!50}87.2} & {\scriptsize 1458.3} & {\scriptsize \cellcolor{c4!15}63.8} & {\scriptsize \cellcolor{c4!15}56.9} & {\scriptsize \cellcolor{c4!50}69.9} & {\scriptsize 65.1} & {\scriptsize \cellcolor{c4!50}58.0} & {\scriptsize \cellcolor{c4!15}17.9} & {\scriptsize \cellcolor{c4!50}24.6} & \cellcolor{gray!25}{\scriptsize \textbf{96.8}} \\
                 \cellcolor{green!10} 23\ \ \ Accuracy as Obj. & {\scriptsize \cellcolor{c4!15}76.9} & {\scriptsize 58.9} & {\scriptsize \cellcolor{c4!15}53.0} & {\scriptsize \cellcolor{c4!15}70.1} & {\scriptsize 57.5} & {\scriptsize \cellcolor{c4!15}87.1} & {\scriptsize \cellcolor{c4!15}1497.6} & {\scriptsize \cellcolor{c4!50}63.9} & {\scriptsize \cellcolor{c4!50}57.6} & {\scriptsize 67.3} & {\scriptsize \cellcolor{c4!15}65.4} & {\scriptsize 57.7} & {\scriptsize \cellcolor{c4!50}18.0} & {\scriptsize \cellcolor{c4!15}24.5} & \cellcolor{gray!25}{\scriptsize \underline{96.5}}\\
             \bottomrule
  
        \end{tabular}
    }
    }
    \label{tab:llava_eval}
    % \vspace{-5mm}
\end{table*}
\footnotetext{Reproduced with official code.}

% \vspace{-1.5mm}
\section{Experiments and Results}
\label{sec:exp}

% \vspace{-1mm}
\subsection{Experimental Setup}
\label{sec:exp_detail}

% \vspace{-1mm}
\textbf{Datasets and Models.}
To demonstrate effectiveness and generalizability across different scales of instruction-tuning (IT) data, we follow \textit{\textbf{standard protocol}} \citep{coincide} \& conduct experiments on two IT datasets: large-scale LLaVA-665K~\citep{llava} containing \textbf{$\sim0.6$ Million samples} \& Vision-Flan-191K~\citep{vision-flan}.
For target VLMs, we primarily use \textbf{LLaVA-v1.5-7B}~\citep{llava}, following prior work, and additionally report results on \textbf{LLaVA-v1.5-13B}~\citep{llava} to test scalability, and \textbf{Qwen2-VL-7B}~\citep{Qwen2VL} and \textbf{Qwen2.5-VL-32B-Instruct}~\citep{bai2025qwen2} to test generalization to newer architectures.

\noindent
\textbf{Implementation Details.} 
{Following standard protocol~\citep{coincide,icons}, we start from a pretrained model and instruction-tune it on the datasets using LoRA~\citep{hu2021lora} with the official hyperparameters from LLaVA-1.5 and Qwen2-VL. For the accuracy variant, cluster-wise accuracy is estimated with an LLM judge that compares VLM outputs to ground-truth answers, though this step is not required for our loss-based variant. We \textbf{\textit{strictly follows}} \textbf{\textit{standard}} setup, evaluation protocols and metrics from prior work to ensure consistency and fair comparison \citep{coincide,icons}. Additional implementation details are provided in {Appendix~\ref{app:impl}}.}

\noindent
\textbf{Baselines.}
Consistent with previous work ~\citep{coincide}, we compare \ours against strong baselines spanning five major categories: (1) scoring function methods (CLIP-Score, EL2N\citep{el2n}, Perplexity~\citep{perplexity}); (2) deduplication-based selection (SemDeDup~\citep{semdedup}); (3) graph-based methods (D2-Pruning~\citep{d2-prune}); (4) external curriculum driven (Self-Filter~\citep{Chen2024selffilter}); and (5) concept-diversity approaches (COINCIDE~\citep{coincide}, Self-Sup~\citep{self-sup}). { Our baselines span both active learning (denoted by \textbf{AL}) and curriculum learning methods (denoted by \textbf{CL}) for comprehensive comparison}. Following previous work, we include \textit{Random}—a competitive baseline shown to perform well due to its diversity—and \textit{Full-Finetune}, representing the performance upper bound with full data/supervision training. Details for baselines follow previous work \citep{coincide} (See {Appendix~\ref{app:baselines}} for further details).
\noindent
\textbf{Evaluation Benchmark.}
Following prior work,
we evaluate our approach on a diverse suite of 14  benchmarks with wide coverage of skills: perceptual reasoning (VQAv2~\citep{balanced_vqa_v2},  VizWiz \citep{gurari2018vizwiz}), textual reasoning (TextVQA \citep{singh2019towards}), compositional reasoning (GQA \citep{hudson2019gqa}), object hallucinations (POPE \citep{pope}), multilingual understanding (MMBench-cn \citep {liu2024mmbench}, CMMMU \citep{zhang2024cmmmu}), instruction-following (LLaVA-Bench\citep{llava}), fine-grained skills (MME \citep{liang2024survey}, MMBench-en \citep{liu2024mmbench}, SEED \citep{li2023seed}), and scientific questions and diagrams (SQA-I \citep{lu2022learn}, AI2D \citep{kembhavi2016diagram}, ChartQA \citep{masry2022chartqa}).

\begin{table}[t]
    \tiny
    \caption{\textbf{Architecture and Dataset Generalization.} 
    For Architecture Generalization, we report Qwen2-VL-7B on the LLaVA-665K dataset using 20\% sampling ratio.
    For Dataset Generalization, we report LLaVA-v1.5-7B on Vision-Flan dataset using 16.7\% sampling ratio following prior work.
    }
    \centering
    {
        \renewcommand{\arraystretch}{1.2}
        \renewcommand{\tabcolsep}{2.5pt}
         \adjustbox{max width=\columnwidth}{
        \begin{tabular}{l c c c c c c c cc c c c}
             \toprule
             {\textbf{Method}} & {\textbf{VQAv2}} & {\textbf{GQA}} & {\textbf{VizWiz}} & {\textbf{SQA-I}} & {\textbf{TextVQA}} & {\textbf{POPE}} & {\textbf{MME}} & \multicolumn{2}{c}{\textbf{MMBench}} & {\textbf{LLaVA-}} & {\textbf{SEED}} & \cellcolor{gray!25} {\textbf{Rel. (\%)}}\\
             & & & & & & & & {\textbf{en}} & {\textbf{cn}} & {\textbf{Wild}} & \\
             \midrule
             \multicolumn{12}{c}{\textbf{Architecture Generalization  (Qwen2-VL-7B)}} \\
             \midrule
             Full-Finetune &
             {\scriptsize 77.4} & {\scriptsize 61.7} & {\scriptsize 45.5} & {\scriptsize 81.4} & {\scriptsize 59.7} & {\scriptsize 84.3} & {\scriptsize 1567.9} & {\scriptsize 76.1} & {\scriptsize 75.1} & {\scriptsize 84.8} & {\scriptsize 66.9} & \cellcolor{gray!25}{\scriptsize 100}\\
             \cmidrule{0-12}
              \cellcolor{green!10}Random &
             {\scriptsize \cellcolor{c4!15}76.2} & {\scriptsize 60.1} & {\scriptsize 43.6} & {\scriptsize 81.4} & {\scriptsize \cellcolor{c4!15}58.7} & {\scriptsize \cellcolor{c4!15}83.7} & {\scriptsize 1556.8} & {\scriptsize 76.8} & {\scriptsize \cellcolor{c4!15}74.5} & {\scriptsize \cellcolor{c4!15}81.7} & {\scriptsize \cellcolor{c4!15}67.6} & \cellcolor{gray!25}{\scriptsize98.7}\\
        
              \cellcolor{lightorange!40}COINCIDE  &
              {\scriptsize  \cellcolor{c4!50}76.7} & {\scriptsize \cellcolor{c4!15}60.2} & {\scriptsize \cellcolor{c4!15}45.4} & {\scriptsize \cellcolor{c4!15}81.7} & {\scriptsize  \cellcolor{c4!50}59.4} & {\scriptsize \cellcolor{c4!15}83.6} & {\scriptsize  \cellcolor{c4!50}1583.5} & {\scriptsize  \cellcolor{c4!50}77.4} & {\scriptsize  \cellcolor{c4!50}76.2} & {\scriptsize 80.5} & {\scriptsize \cellcolor{c4!50}67.9} & \cellcolor{gray!25}{\scriptsize \underline{99.6}}\\

             \cellcolor{green!10}\textbf{\ours}  & {\scriptsize \cellcolor{c4!15}76.2} & {\scriptsize  \cellcolor{c4!50}60.5} & {\scriptsize  \cellcolor{c4!50}47.1} & {\scriptsize  \cellcolor{c4!50}82.3} & {\scriptsize 58.0} & {\scriptsize  \cellcolor{c4!50}84.3} & {\scriptsize \cellcolor{c4!15}1560.1} & {\scriptsize \cellcolor{c4!15}77.2} & {\scriptsize 72.9} & {\scriptsize  \cellcolor{c4!50}87.1} & {\scriptsize \cellcolor{c4!15}67.6} & \cellcolor{gray!25}{\scriptsize \textbf{100.0}}\\
             \midrule
             \multicolumn{12}{c}{\textbf{Dataset Generalization (Vision-Flan-191K)}} \\
             \midrule
             Full-Finetune &
             {\scriptsize 69.4} & {\scriptsize 46.0} & {\scriptsize 49.7} & {\scriptsize 59.9} & {\scriptsize 34.1} & {\scriptsize 85.1} & {\scriptsize 1306.1} & {\scriptsize 49.1} & {\scriptsize 51.7} & {\scriptsize 35.7} & {\scriptsize 53.3} & \cellcolor{gray!25}{\scriptsize 100}\\
             \cmidrule{0-12}
              \cellcolor{green!10}Random &
              {\cellcolor{c4!15}\scriptsize 66.0} & {\scriptsize \cellcolor{c4!15}43.8} & {\scriptsize \cellcolor{c4!15}52.2} & {\scriptsize 62.1} & {\scriptsize \cellcolor{c4!15}39.7} & {\scriptsize \cellcolor{c4!15}82.7} & {\scriptsize \cellcolor{c4!15}1072.2} & {\cellcolor{c4!15}\scriptsize 48.7} & {\scriptsize 43.7} & {\scriptsize \cellcolor{c4!15}40.4} & {\scriptsize 28.7}& \cellcolor{gray!25}{\scriptsize {95.0}}\\
        
              \cellcolor{lightorange!40}COINCIDE  &
               {\scriptsize \cellcolor{c4!50}66.3} & {\scriptsize 43.6} & {\scriptsize 51.0} & {\cellcolor{c4!50}\scriptsize 63.8} & {\scriptsize 35.2} & {\scriptsize 81.9} & {\cellcolor{c4!50}\scriptsize 1222.2} & {\cellcolor{c4!50}\scriptsize 56.7} & {\scriptsize \cellcolor{c4!15}45.5} & {\scriptsize 31.1} & {\scriptsize \cellcolor{c4!15}37.5} & \cellcolor{gray!25}{\scriptsize \underline{95.8}}\\

             \cellcolor{green!10}\textbf{\ours}  & {\scriptsize 65.5} & {\cellcolor{c4!50}\scriptsize 44.0} & {\cellcolor{c4!50}\scriptsize 53.6} & {\scriptsize \cellcolor{c4!15}62.5} & {\cellcolor{c4!50}\scriptsize 42.0} & {\cellcolor{c4!50}\scriptsize 82.9} & {\scriptsize 1040.9} & {\scriptsize 43.6} & {\cellcolor{c4!50}\scriptsize 47.4} & {\cellcolor{c4!50}\scriptsize 43.2} & {\scriptsize \cellcolor{c4!50}45.3}& \cellcolor{gray!25}{\scriptsize \textbf{99.0}}\\
             \bottomrule
  
        \end{tabular}
        }
    }
    \label{tab:generalization}
    % \vspace{-5mm}
\end{table}

\noindent
\textbf{Evaluation Metrics.}
Following all prior work, we use \textbf{\textit{standard evaluation metrics}} to ensure consistency 
and fair comparison. 
Specifically, we report \textbf{average relative performance} \citep{coincide} across benchmarks to provide a unified measure of generalization. For each benchmark, relative performance is defined as:
$\text{Rel.} = \left(\frac{\text{Model Performance}}{\text{Full Data Finetuned Performance}}\right) \times 100\%$. This normalizes differences in performance scale and difficulty across benchmarks, following prior work.

% \vspace{-1mm}
\subsection{Results and Analysis}
\label{results}
% \vspace{-2.2pt}
\textbf{\ours is more effective than existing SOTA in data efficient learning.}
Table~\ref{tab:llava_eval} (Row 0-11) compares \ours against state-of-the-art baselines for training \textbf{LLaVA-v1.5-7B} on LLaVA-665K dataset under a 20\% data budget, following \textbf{\textit{standard protocol}}. \ours achieves the highest relative performance (98.8\%), outperforming all baselines, including those requiring access to ground-truth answers for the entire dataset and additional reference VLMs. In contrast, \ours uses supervision only on a \emph{need basis} for 20\% of samples and relies solely on model's own progress, yet reaching near-parity with full finetuning. Beyond aggregate gains, \ours also ranks among the top two methods on 8 out of 14 benchmarks, showing strong generalization across diverse tasks (Table~\ref{tab:llava_eval} (Row 0-11))—{\em e.g.}, including perception-focussed VQAv2 (75.2), scientific questions and diagrams (ChartQA:17.3, AI2D:52.8), and object hallucination POPE (85.9). Notably, it exceeds full-data performance on VizWiz (53.4 vs. 47.8), SQA-I (69.9 vs. 68.4), 
% POPE (86.3 vs. 86.4), 
MME (1483.2 vs. 1476.9), 
ChartQA (17.3 vs. 16.4)
and CMMMU (24.6 vs. 22.1). These results demonstrate \ours is a dynamic and fully automated alternative for efficient VLM training.

% \vspace{-2pt}
\textbf{Scalability to Larger Models.}
To assess scalability, we use \ours to train the larger \textbf{LLaVA-v1.5-13B} model under the same 20\% data budget, testing whether our method developed for LLaVA-v1.5-7B transfers effectively to a higher-capacity model without hyperparameter tuning.
As shown in Table~\ref{tab:llava_eval} (Row 12-23), \ours achieves a relative performance of 96.8\%, outperforming all baselines.
Beyond aggregate gains, \ours ranks among the top-2 methods on 8 out of 14 benchmarks compared with all baselines, demonstrating strong generalization.

% \vspace{-2pt}
\textbf{VLM Architectures and Dataset Generalization.}
In Table~\ref{tab:generalization}, we test  generalization of PROGRESS across different VLM architecture and IT dataset with accuracy as signal.
For \textbf{architecture generalization}, we use newer \textbf{Qwen2-VL-7B} and train it on the LLaVA-665K dataset using the same 20\% data budget and identical hyperparameters.
We compare \ours with two of the strongest (highest performing)  established baselines—Random Sampling and COINCIDE—across multiple multimodal benchmarks.
\ours achieves the highest overall relative performance of 100\% and ranks first or second on 9 out of 11 benchmarks (Tab.~\ref{tab:generalization}, \textbf{top}).
{We also report results on \textbf{Qwen2.5-VL-32B-Instruct} (See {Appendix~\ref{app:qwen_32b}}), demonstrating the strong generalization to even larger-scale VLMs.}
For \textbf{dataset generalization}, we report LLaVA-v1.5-7B on Vision-Flan dataset under a stricter 16.7\% annotation budget (\textbf{\textit{standard protocol}}) to assess generalization in low-resource settings. 
\ours achieves the highest overall relative performance of 99.0\%, outperforming COINCIDE (95.8\%) and Random (95.0\%) and ranks first or second on 8 out of 11 benchmarks (Tab.~\ref{tab:generalization}, \textbf{bottom}).
These results underscore the calability and generalization of \ours, making it a practical solution for efficient training across diverse architecture and datasets.

% \vspace{-1mm}
\subsection{Investigating the effectiveness of different components of PROGRESS}
\label{sec:abl}

% \vspace{-2mm}
We analyze \& ablate the components of \ours. We use LLaVA-v1.5-7B on LLaVA-665K with 20\% sampling and accuracy as the objective unless stated otherwise.

\begin{table}[t]
    \tiny
    \caption{\textbf{Ablation of Selection Policy.} Performance comparison of \ours with baselines and curriculum learning (CL)-based selection strategies under identical warmup conditions.
    }
    \centering
    {
        \renewcommand{\arraystretch}{1.25}
        \renewcommand{\tabcolsep}{1.4pt}
         \adjustbox{max width=\columnwidth}{
        \begin{tabular}{l c c c c c c c cc c c c c c c}
             \toprule
             {\textbf{Method}} & {\textbf{VQAv2}} & {\textbf{GQA}} & {\textbf{VizWiz}} & {\textbf{SQA-I}} & {\textbf{TextVQA}} & {\textbf{POPE}} & {\textbf{MME}} & \multicolumn{2}{c}{\textbf{MMBench}} & {\textbf{LLaVA-}}  &{\textbf{SEED}}  & {\textbf{AI2D}}   & {\textbf{ChartQA}} & {\textbf{CMMMU}}  &  \cellcolor{gray!25}{\textbf{Rel. (\%)}}\\
             & & & & & & & & {\textbf{en}} & {\textbf{cn}} & {\textbf{Wild}} &  & \\
             \midrule
             \ 0\ \ \ Full-Finetune &
             {\scriptsize 79.1} & {\scriptsize 63.0} & {\scriptsize 47.8} & {\scriptsize 68.4} & {\scriptsize 58.2} & {\scriptsize 86.4} & {\scriptsize 1476.9} & {\scriptsize 66.1} & {\scriptsize 58.9} & {\scriptsize 67.9}  & {\scriptsize 67.0} & {\scriptsize 56.4} & {\scriptsize 16.4} & {\scriptsize 22.1} & \cellcolor{gray!25}{\scriptsize 100}\\
             \cmidrule{0-15}
             \cmidrule{0-15}
            \ 1\ \ \ Random & {\scriptsize \cellcolor{c4!15}75.7} & {\scriptsize \cellcolor{c4!50}59.0} & {\scriptsize 43.8} & {\scriptsize 68.8} & {\scriptsize 54.9} & {\scriptsize 85.6} & {\scriptsize 1414.2} & {\scriptsize 61.9} & {\scriptsize \cellcolor{c4!15}54.9} & {\scriptsize 66.2} & {\scriptsize \cellcolor{c4!15}63.3} & {\scriptsize 48.6} & {\scriptsize \cellcolor{c4!15}17.3} & {\scriptsize 25.2} & \cellcolor{gray!25}{\scriptsize {96.8}}\\
            \makecell[l]{\ 2\ \ \ COINCIDE  \\ \hspace{1.2em} \ [\textit{{DINO-BERT}}]}  & {\scriptsize \cellcolor{c4!50}76.0} & {\scriptsize 58.3} & {\scriptsize 40.1} & {\scriptsize 67.8} & {\scriptsize \cellcolor{c4!50}55.7} & {\scriptsize \cellcolor{c4!50}87.2} & {\scriptsize 1466.1} & {\scriptsize \cellcolor{c4!15}62.2} & {\scriptsize 53.8} & {\scriptsize \cellcolor{c4!50}69.1} & {\scriptsize \cellcolor{c4!15}63.3} & {\scriptsize 52.6} & {\scriptsize \cellcolor{c4!50}17.6} & {\scriptsize 23.7} & \cellcolor{gray!25}{\scriptsize {96.9}} \\
             \ 3\ \ \ Easiest (\textbf{\textit{CL}})  & {\scriptsize 72.0} & {\scriptsize 54.8} & {\scriptsize 50.2} & {\scriptsize 67.1} & {\scriptsize 51.6} & {\scriptsize 85.7} & {\scriptsize 1407.4} & {\scriptsize 57.0} & {\scriptsize 52.6} & {\scriptsize 65.2} & {\scriptsize 59.5} & {\scriptsize 50.1} & {\scriptsize 12.3} & {\scriptsize 22.8} & \cellcolor{gray!25}{\scriptsize 92.3}\\
             \ 4\ \ \ Medium (\textbf{\textit{CL}}) & {\scriptsize 69.3} & {\scriptsize 52.5} & {\scriptsize 46.0} & {\scriptsize 68.3} & {\scriptsize 50.8} & {\scriptsize 85.4} & {\scriptsize 1307.6} & {\scriptsize 54.6} & {\scriptsize 48.7} & {\scriptsize 62.5} & {\scriptsize 57.7} & {\scriptsize 47.6} & {\scriptsize 14.3} & {\scriptsize \cellcolor{c4!50}26.1} & \cellcolor{gray!25}{\scriptsize 91.1}\\
             \ 5\ \ \ Hardest (\textbf{\textit{CL}}) & {\scriptsize 72.8} & {\scriptsize 54.8} & {\scriptsize \cellcolor{c4!15}52.1} & {\scriptsize 61.3} & {\scriptsize 50.5} & {\scriptsize 85.4} & {\scriptsize 1364.8} & {\scriptsize 37.9} & {\scriptsize 34.5} & {\scriptsize \cellcolor{c4!15}67.5} & {\scriptsize 54.1} & {\scriptsize 41.4} & {\scriptsize 15.8} & {\scriptsize \cellcolor{c4!15}25.9} & \cellcolor{gray!25}{\scriptsize 88.5}\\
             
             \textbf{\ours } \\ 
              \ 6\ \ \ Loss as Obj.  & {\scriptsize \cellcolor{c4!15}75.7} & {\scriptsize 58.6} & {\scriptsize 49.6} & {\scriptsize \cellcolor{c4!50}70.1} & {\scriptsize \cellcolor{c4!15}55.1} & {\scriptsize \cellcolor{c4!15}86.3} & {\scriptsize \cellcolor{c4!50}1498.4} & {\scriptsize \cellcolor{c4!50}62.5} & {\scriptsize \cellcolor{c4!50}55.5} & {\scriptsize 65.5} & {\scriptsize \cellcolor{c4!50}63.4} & {\scriptsize \cellcolor{c4!50}53.3} & {\scriptsize \cellcolor{c4!15}17.3} & {\scriptsize 23.7} & \cellcolor{gray!25}{\scriptsize \underline{98.4}} \\
              \ 7\ \ \  Accuracy as Obj. & {\scriptsize 75.2} & {\scriptsize \cellcolor{c4!15}58.8} & \cellcolor{c4!50}{\scriptsize 53.4} & \cellcolor{c4!15}{\scriptsize 69.9} & {\scriptsize \cellcolor{c4!15}55.1} & {\scriptsize 85.9} & {\scriptsize \cellcolor{c4!15}1483.2} & {\scriptsize 61.1} & {\scriptsize 54.4} & {\scriptsize 65.5} & {\scriptsize 63.0} & {\scriptsize \cellcolor{c4!15}52.8} & {\scriptsize  \cellcolor{c4!15}17.3}  & {\scriptsize  24.6} & \cellcolor{gray!25}{\scriptsize \textbf{98.8}} \\
             \bottomrule
  
        \end{tabular}
        }
    }
    \label{tab:ablation_policy}
    % \vspace{-5mm}
\end{table}

\begin{figure}[t]
% \vspace{-0.3in}
    \begin{center}
    \begin{subfigure}[t]{0.32\linewidth}
        \includegraphics[width=\linewidth]{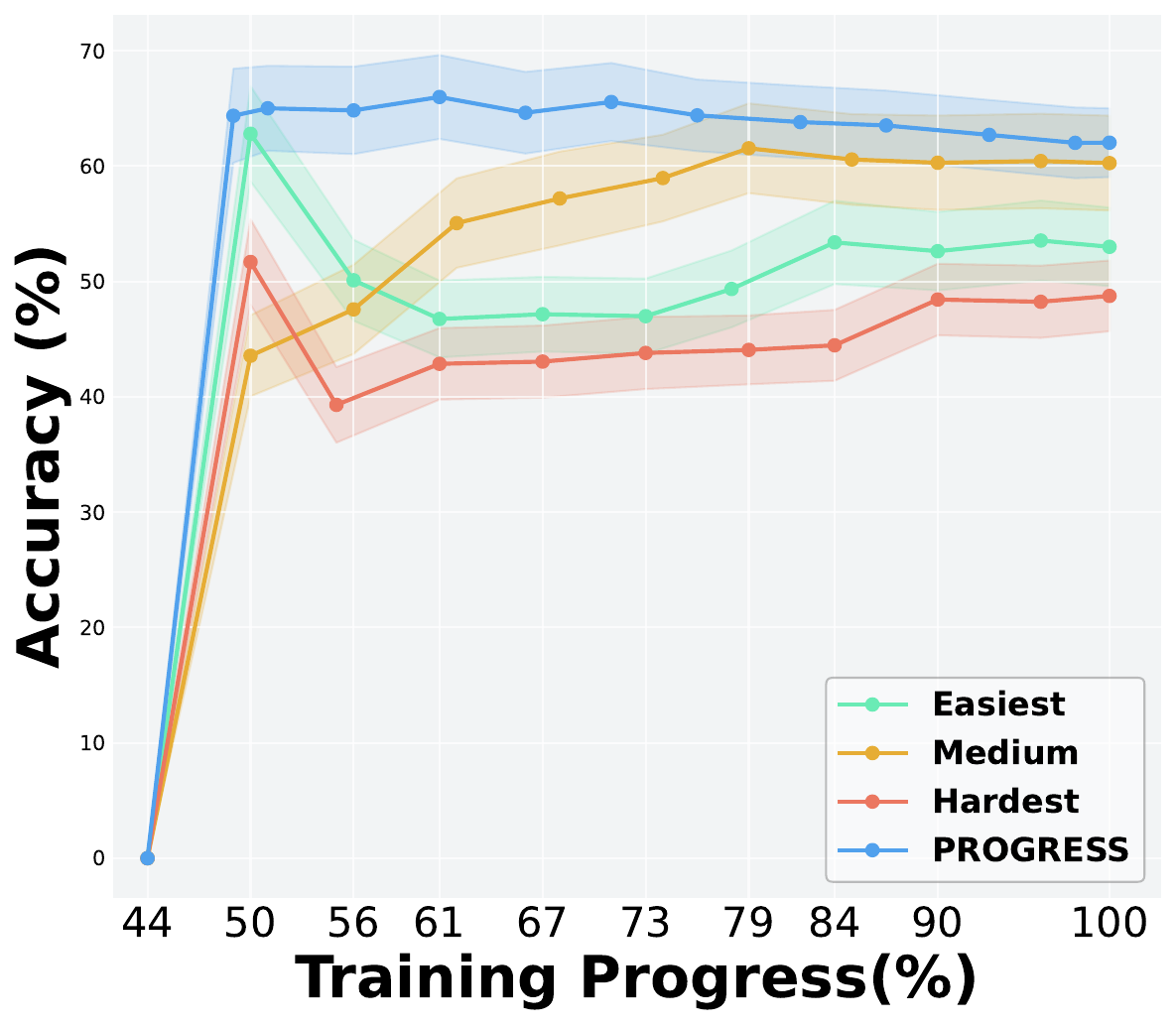}
        \caption{Hard Skills}
        \label{fig:cur_hardest}
    \end{subfigure}
    \begin{subfigure}[t]{0.32\linewidth}
        \includegraphics[width=\linewidth]{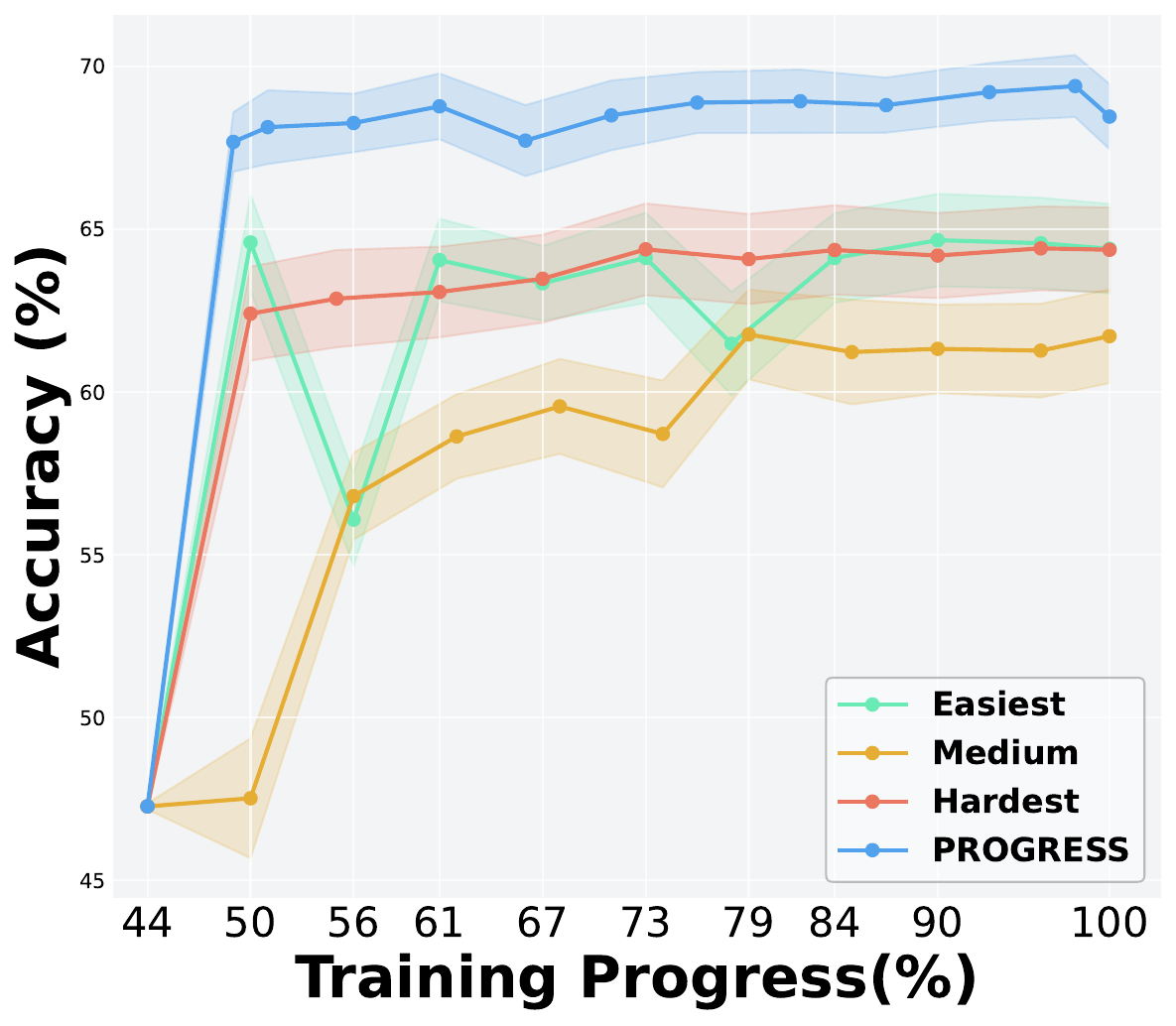}
        \caption{Moderate Skills}
        \label{fig:cur_mid} 
    \end{subfigure}
    \begin{subfigure}[t]{0.32\linewidth}
        \includegraphics[width=\linewidth]{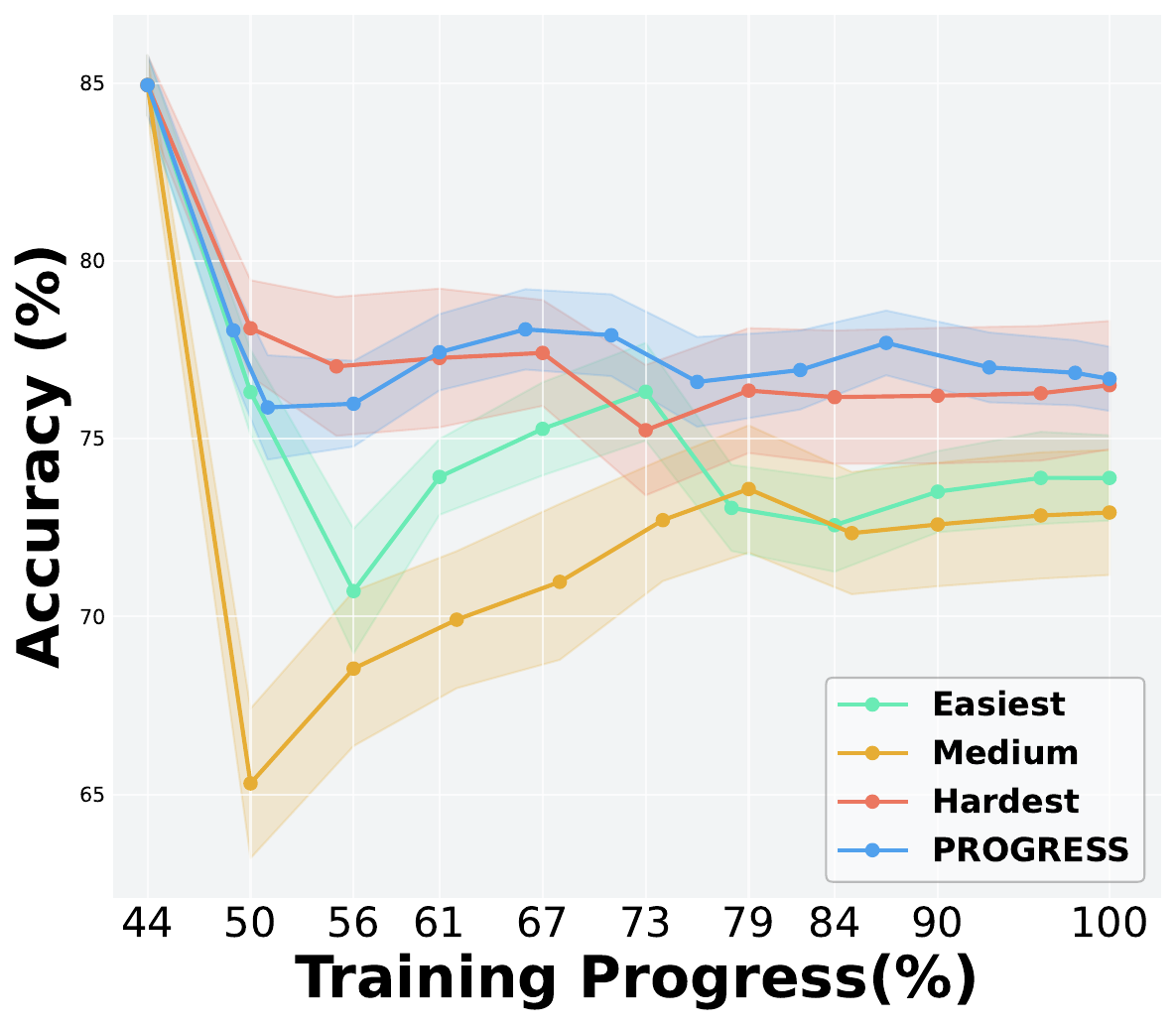}
        \caption{Easy Skills}
        \label{fig:cur_easiest}
    \end{subfigure}
    \caption{\textbf{Learning Dynamics Across Difficulty Levels.} \ours consistently achieves higher accuracy and reduced variance compared to other selection strategies}
    \label{fig:cur_easy_mid_hard}
    \end{center}
    \vspace{-0.35in}
\end{figure}

% \noindent
\textbf{How effective is our Selection Policy (i.e Eqn \ref{eq:delta})? }
We evaluate our progress-based selection policy (Sec.~\ref{PCL}) against several competitive strategies in Table~\ref{tab:ablation_policy}.
All strategies use identical warmup conditions.
We compare with curriculum baselines: \textbf{Random Selection}, \textbf{Easiest-Sampling} (selecting clusters with highest absolute accuracy at a given step), \textbf{Medium-Sampling} (selecting mid-accuracy clusters), and \textbf{Hardest-Sampling} (selecting lowest-accuracy clusters).
As shown in Tab. \ref{tab:ablation_policy}, \ours achieves the highest relative score (98.8\%), ranking first on 7 out of 14 benchmarks and second on 5 others.
To further analyze learning dynamics across strategies, we group skill clusters into difficulty levels: easy, moderate, and hard based on their initial performance, and track the average performance of 50 clusters per level (see Fig.~\ref{fig:cur_easy_mid_hard}). \ours consistently achieves higher mean performance with lower variance across all levels, effectively balancing learning across task difficulties throughout training. Row~2 includes a \textbf{DINO–BERT version of COINCIDE}, which replaces Tiny-LLaVA features with the same unsupervised features used in our method, confirming that the gains of \ours arise from the \textbf{progress-driven selection policy} rather than the warmup or feature choice.

\setlength{\intextsep}{8pt}  % 默认是 12pt，调小可以减少上下空白
\setlength{\columnsep}{9pt}  % 调整图片与文字之间的间距

\begin{wrapfigure}{r}{0.23\textwidth}
% \vspace{-0.3in}
    \centering
    \includegraphics[width=\linewidth]{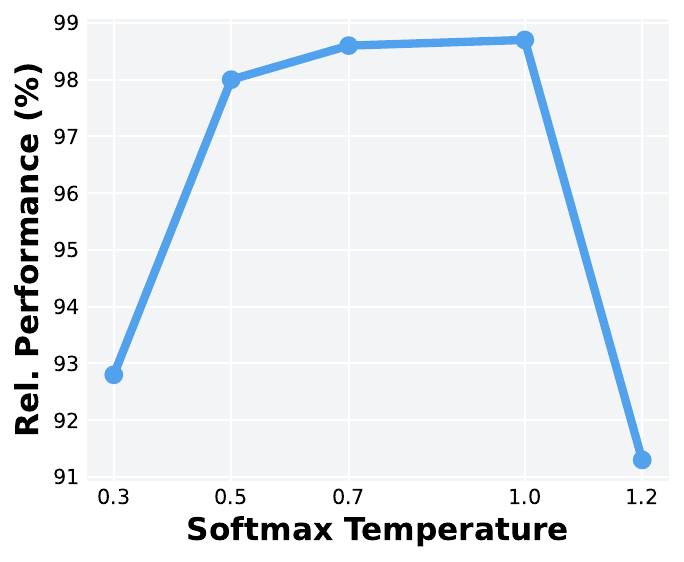}
    \caption{\textbf{Ablation of Softmax Temperature.} Both very-low and very-high temperatures lead to a significant performance drop.}
    \label{fig:abl_temp}
% \vspace{-2.5mm}
\end{wrapfigure}

% \noindent
\textbf{How important is balancing informativeness \& diversity in selected samples (i.e Eqn \ref{eq:softmax})?}
We ablate the \textit{temperature} $\tau$ in the softmax skill-selection (Eqn.~\ref{eq:softmax}). As discussed in Sec \ref{PCL}, lower $\tau$ would over-focuses on top clusters, reducing diversity, while very high $\tau$ makes high-improvement clusters lose priority. Fig.~\ref{fig:abl_temp} shows, $\tau=1.0$ achieves the best performance (98.8\% Rel.), balancing priority and diversity. Decreasing $\tau$ (0.7, 0.5, 0.3) degrades performance, with lowest $\tau=0.3$ dropping to 92.8\% (-6\%), confirming that overly sharp distributions cause \textit{mode collapse}. Excessive diversity ($\tau=1.2$) also hurts performance, as high-improvement clusters start to lose their clear priority during selection.

% \noindent
\textbf{Ablation of Warmup}.
{
Table \ref{tab:ablation_warmup} shows that warmup phase alone provides only modest performance (94.6\% Rel.), while \ours yields large gains (+4\% relative) over the warmup-only baseline (rows 2 vs. row 1). This demonstrates that warmup merely initializes reasonable per-cluster estimates, whereas the major improvements stem from \ours's progress-driven sample selection. Row 3 shows \ours remains robust to different warmup strategies, consistently delivering near-full-data performance even with weak (random sampling) warmup.
}

\begin{table}[t]
    \tiny
    \caption{\textbf{Ablation of Warmup.} Performance comparison of \ours against warmup-only and across different warmup strategies. \ours significantly boosts performance over warmup and remains robust across different warmup strategies.}
    \centering
    {
        \renewcommand{\arraystretch}{1.25}
        \renewcommand{\tabcolsep}{1.4pt}
         \adjustbox{max width=\columnwidth}{
        \begin{tabular}{l c c c c c c c cc c c c c c c}
             \toprule
             {\textbf{Method}} & {\textbf{VQAv2}} & {\textbf{GQA}} & {\textbf{VizWiz}} & {\textbf{SQA-I}} & {\textbf{TextVQA}} & {\textbf{POPE}} & {\textbf{MME}} & \multicolumn{2}{c}{\textbf{MMBench}} & {\textbf{LLaVA-}}  &{\textbf{SEED}}  & {\textbf{AI2D}}   & {\textbf{ChartQA}} & {\textbf{CMMMU}}  &  \cellcolor{gray!25}{\textbf{Rel. (\%)}}\\
             & & & & & & & & {\textbf{en}} & {\textbf{cn}} & {\textbf{Wild}} &  & \\
             \midrule
             \ 0\ \ \ Full-Finetune &
             {\scriptsize 79.1} & {\scriptsize 63.0} & {\scriptsize 47.8} & {\scriptsize 68.4} & {\scriptsize 58.2} & {\scriptsize 86.4} & {\scriptsize 1476.9} & {\scriptsize 66.1} & {\scriptsize 58.9} & {\scriptsize 67.9}  & {\scriptsize 67.0} & {\scriptsize 56.4} & {\scriptsize 16.4} & {\scriptsize 22.1} & \cellcolor{gray!25}{\scriptsize 100}\\
             \cmidrule{0-15}
             \cmidrule{0-15}
             \ 1\ \ \ Warmup Only  & {\scriptsize 73.1} & {\scriptsize 55.9} & {\scriptsize 43.8} & {\scriptsize \cellcolor{c4!15}67.9} & {\scriptsize 54.2} & {\scriptsize \cellcolor{c4!15}85.4} & {\scriptsize 1410.3} & {\scriptsize 58.5} & {\scriptsize 52.7} & {\scriptsize 64.6} & {\scriptsize 60.5} & {\scriptsize 52.4} & {\scriptsize 16.1} & {\scriptsize \cellcolor{c4!15}24.5} & \cellcolor{gray!25}{\scriptsize 94.6} \\
               \makecell[l]{\ 2\ \ \ Warmup   \\ 
               \hspace{1.0em} \ + PROGRESS}  & {\scriptsize \cellcolor{c4!50}75.2} & {\scriptsize \cellcolor{c4!50}58.8} & \cellcolor{c4!50}{\scriptsize 53.4} & \cellcolor{c4!50}{\scriptsize 69.9} & {\scriptsize \cellcolor{c4!50}55.1} & {\scriptsize \cellcolor{c4!50}85.9} & {\scriptsize \cellcolor{c4!50}1483.2} & {\scriptsize \cellcolor{c4!50}61.1} & {\scriptsize \cellcolor{c4!50}54.4} & {\scriptsize \cellcolor{c4!15}65.5} & {\scriptsize \cellcolor{c4!50}63.0} & {\scriptsize \cellcolor{c4!15}52.8} & {\scriptsize  \cellcolor{c4!15}17.3}  & {\scriptsize  \cellcolor{c4!50}24.6} & \cellcolor{gray!25}{\scriptsize \textbf{98.8}} \\

              \makecell[l]{\ 3\ \ \ Random Warmup   \\ 
               \hspace{1.0em} \ + PROGRESS} & {\scriptsize \cellcolor{c4!15}75.0} & {\scriptsize \cellcolor{c4!15}58.0} & {\scriptsize \cellcolor{c4!15}46.1} & {\scriptsize 67.4} & {\scriptsize \cellcolor{c4!15}54.7} & {\scriptsize 85.0} & {\scriptsize \cellcolor{c4!15}1459.2} & {\scriptsize \cellcolor{c4!15}60.6} & {\scriptsize \cellcolor{c4!15}54.3} & {\scriptsize \cellcolor{c4!50}67.3} & {\scriptsize \cellcolor{c4!15}61.7} & {\scriptsize \cellcolor{c4!50}53.8} & {\scriptsize \cellcolor{c4!50}17.7} & {\scriptsize 23.8} & \cellcolor{gray!25}{\scriptsize \underline{97.1}} \\
              
             \bottomrule
  
        \end{tabular}
        }
    }
    \label{tab:ablation_warmup}
    \vspace{-5mm}
\end{table}

\begin{wrapfigure}{r}{0.23\textwidth}
% \vspace{-0.1in}
    \begin{center}
    \includegraphics[width=0.23\textwidth]{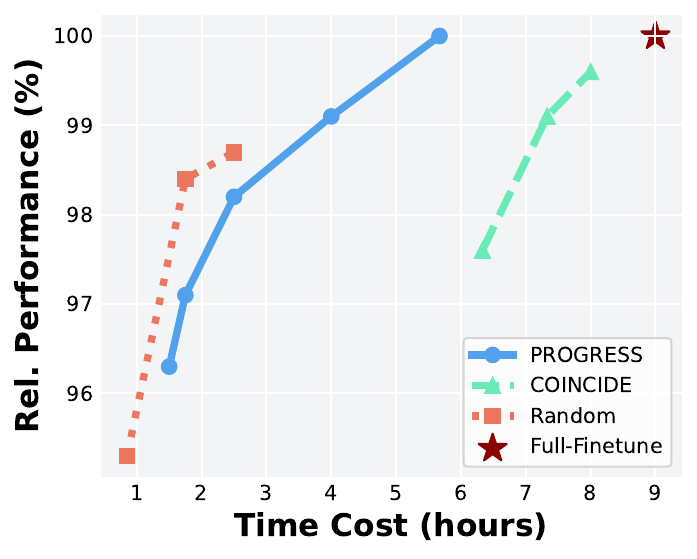}
    \caption{\textbf{Wall-clock time} comparison. {We show results on LLaVA-665K here and Vision-Flan-191K in {App. \ref{app:wall_clock_app}}.}}
    \label{fig:wall_clock_time_main}
    \end{center}
    \vspace{-0.15in}
\end{wrapfigure}

\textbf{Time Efficiency and Annotation Cost Analysis.}
Consistent with prior work~\citep{coincide}, we measure the wall-clock cost of the \textbf{\textit{entire pipeline}}—data selection plus model finetuning—against relative performance (Rel.). 
As shown in Fig. \ref{fig:wall_clock_time_main}, PROGRESS reaches relative performances of 96.3\%, 97.1\%, 98.2\%, 99.1\%, and 100\% within 1.5, 1.75, 2.5, 4, and 5.67 hrs, respectively, making it 
Pareto-superior to COINCIDE. 
Furthermore, Full-data finetuning requires $\sim$9 hr (and 100\% data for training), much higher than our method which needs only 5.67hr of total training time and 20\% data for training. 

\textbf{Note:} Runtime includes all cost for  \textbf{entire
pipeline}, for fair comparison (Fig. \ref{fig:wall_clock_time_main}), specifically: PROGRESS comprises feature extraction, K-means clustering, self-evaluation, warm-up and training, while COINCIDE includes feature extraction, clustering, and training. Both methods use same GPU compute and 20\% data to train following standard protocol from prior work. See {Appendix~\ref{app:time_break_down}} for detailed stage-wise time breakdown.
Beyond wall-clock gains, \textbf{annotation cost} dominates end-to-end cost in VLM training ($\approx$1902 hr for LLaVA-665K); by requiring only 20\% labels, \ours reduces annotation time by 80\% ($\approx$380 hr), yielding the largest overall savings when combining annotation and training.
See Appendix~\ref{app:annotation_time} details.
\begin{wrapfigure}{r}{0.23\textwidth}
% \vspace{-0.1in}
    \begin{center}
    \includegraphics[width=0.23\textwidth]{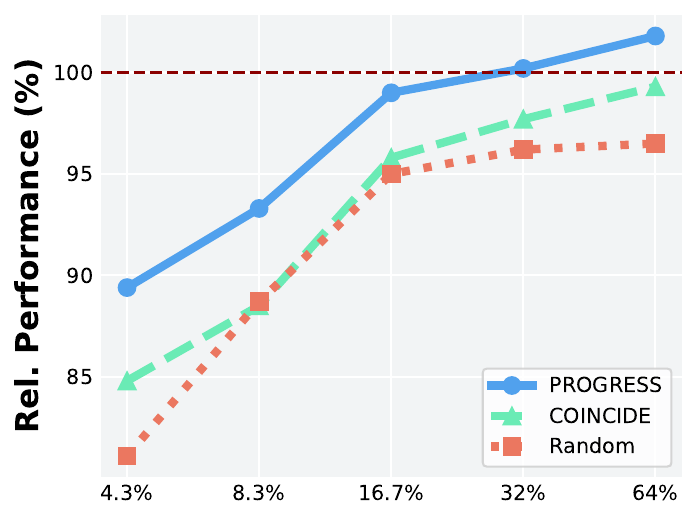}
    \caption{\textbf{Ablation with different data budget for training}. Relative performance on Vision-Flan dataset under different sampling ratio.}
    \label{fig:sample_ratio}
    \end{center}
    \vspace{-0.2in}
\end{wrapfigure}

% \vspace{-0.05in}
\textbf{How effective is \ours under different data budget?} 
Fig~\ref{fig:sample_ratio} shows performance under different 
data budget used for training. 
% (even lower than 16.7 \% considered in Tab. \ref{tab:generalization}) 
\ours consistently outperforms strongest baselines - Random and COINCIDE across different sampling ratios, highlighting its effectiveness. Notably, on scaling data size to higher percentages 32\%, 64\%, our method outperforms full-data-finetuning (which uses 100\% data) by larger and larger margin. See {Appendix \ref{app:scaling_data}} for
more details on how scaling data improves PROGRESS more than Full data-Finetuning.

\textbf{Representativeness of the Selected Subset.}
We assess how well each selected subset represents the full data distribution using PCA-based representativeness scores in the data feature space. \ours yields a substantially lower representativeness distance (lower is better) to the full data pool ($|$0.083$|$ vs. $|$0.198$|$), reflecting stronger global coverage. \ours retains more variance than COINCIDE (38.7\% vs. 34.2\%), and captures a larger fraction of the top principal components (77–88\% vs. 48–72\%). These results show that \ours spans the overall data distribution more broadly, while COINCIDE selects a narrower, more concentrated subset. See Appendix \ref{app:ablation} for analysis details

% \vspace{-0.1in}
\noindent{\textbf{Ablation Sensitivity to Hyperparameters}}:\\ (Clusters K,$\gamma$, warm-up ratio, etc) -
See {Appendix~\ref{app:ablation}}.

% \vspace{-0.05in}
\noindent{\textbf{Importance of skill order?}} See Appendix~\ref{app:ablation}.

\begin{figure}[t]
% \vspace{-0.3in}
    \begin{center}
    \begin{subfigure}[b]{0.21\textwidth}
        \includegraphics[width=\textwidth]{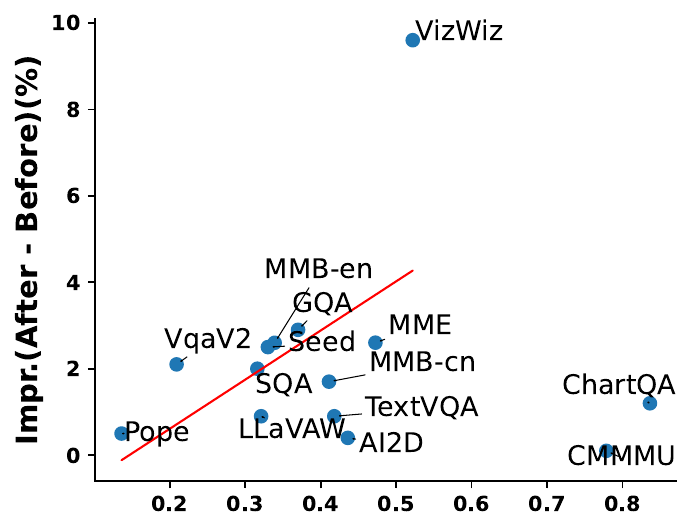}
        \caption{ Difficulty}
        \label{fig:difficulty}
    \end{subfigure}
    \begin{subfigure}[b]{0.21\textwidth}
        \includegraphics[width=\textwidth]{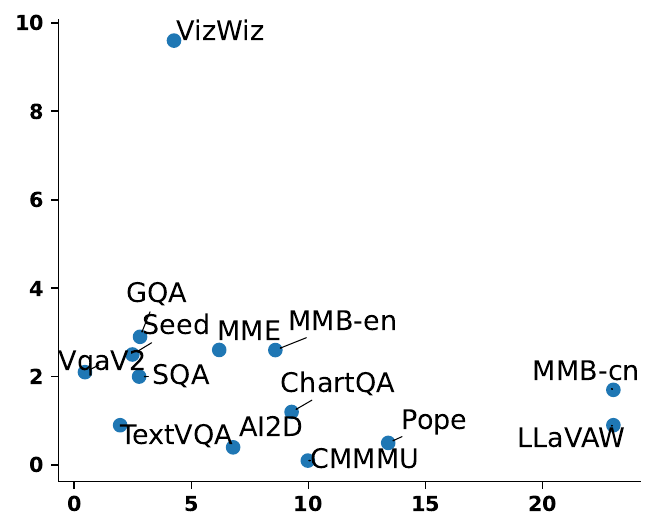}
        \caption{Rarity}
        \label{fig:rarity} 
    \end{subfigure}
    \caption{\textbf{Accuracy improvements} with \textbf{(a) benchmark difficulty} and \textbf{(b) sample rarity}. Largest performance gains occur in mid-range of difficulty \& low-mid range of rarity.}
    \label{fig:difficult_rarity}
    \end{center}
    \vspace{-9mm}
\end{figure}

% \vspace{-0.15in}
% \vspace{-1mm}
\subsection{Analyzing Model Learning Behavior}
\label{sec:analysis}
% \vspace{-1mm}

\textbf{How does benchmark difficulty and data frequency impact performance?}
 % Here, we further examine \textit{how improvements from \ours vary with benchmark difficulty and data frequency}. 
 In Fig.~\ref{fig:difficult_rarity}(a), we plot accuracy gains/improvement of \ours as a function of benchmark difficulty, defined as the gap from full-data finetuned performance ($ (100 - \text{full-finetune score})/100 $) (details in {Appendix~\ref{app:bench_diff}}). \ours achieves the largest gains on benchmarks of moderate difficulty. At the extremes, easy tasks such as POPE show limited improvements due to performance saturation, while hard tasks such as ChartQA, CMMMU yield smaller absolute gains because chart-related skills \& rare Chinese-language skills are underrepresented in LLaVA-665K training dataset($\sim$0.96\% \& $\sim$1.1\% respectively). 
Despite this scarcity, \ours still surpasses full-data finetuning performance on both ChartQA and CMMMU with only 20\% training data (Table~\ref{tab:llava_eval}, rows 10--11), showing that it can \emph{enhance} performance of rare or niche abilities under limited supervision, though scarcity naturally constrains absolute gains. 
Fig.~\ref{fig:difficult_rarity}(b) further confirms this trend by plotting improvements against rarity, measured as $\log(1/\text{frequency})$ of benchmark-aligned samples (details in {Appendix~\ref{app:data_frequency}}): once again, \ours performs best in the mid-rarity regime, prioritizing skills that are neither over-abundant (with limited additional benefit from selecting more samples) nor rare (too few samples to generalize).
Our findings align with the Zone of Proximal Development~\citep{vygotsky2012thought} \& findings in \citep{DataEnvGym}, which state learning is most effective just beyond current ability where learning potential is highest—neither too easy nor too difficult.

\begin{figure}[t]
    \centering
    \begin{subfigure}[t]{0.48\linewidth}
       \includegraphics[width=\textwidth]{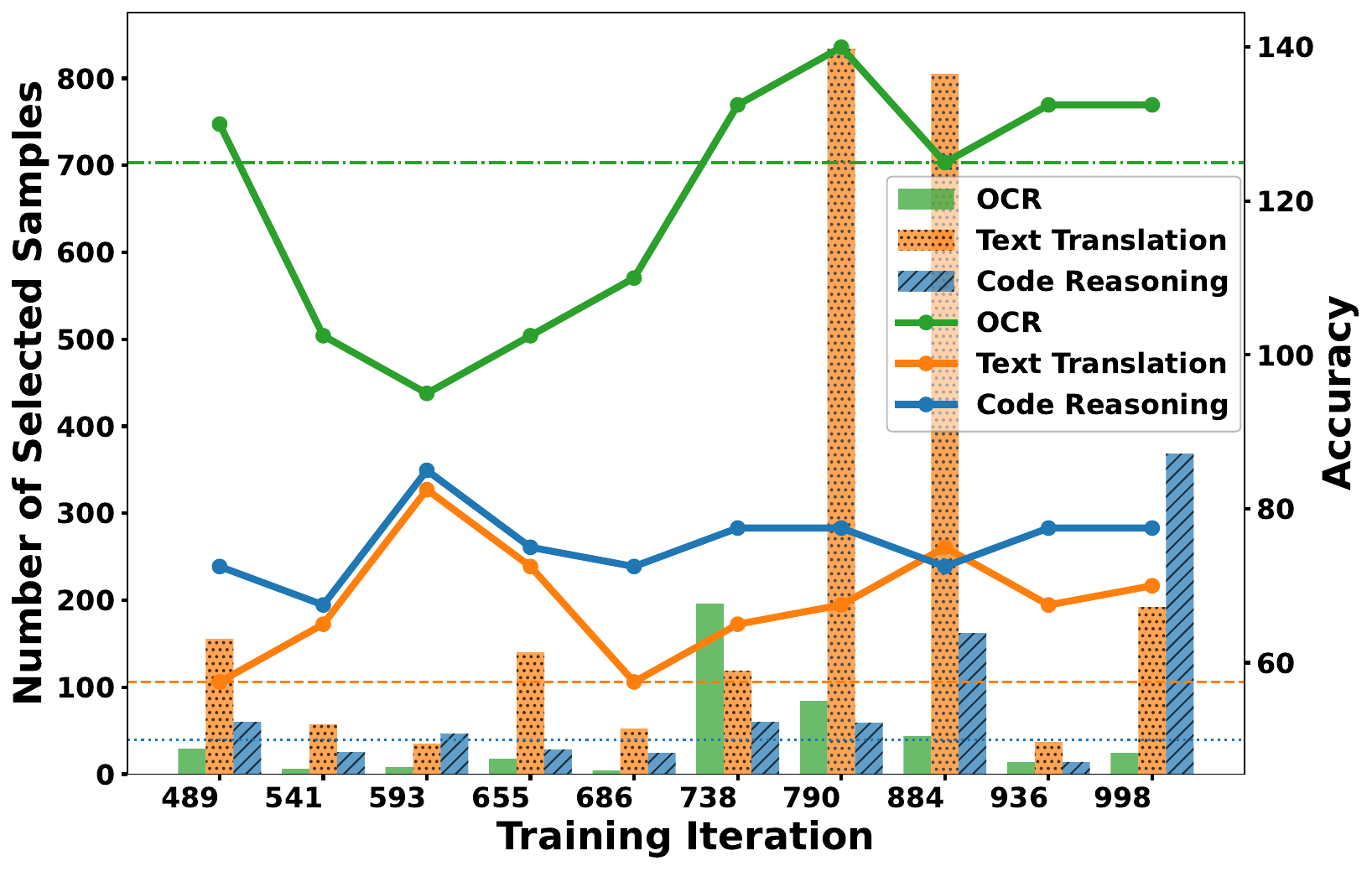}
       \caption{Language Abilities}
       \label{fig:ability_hard}
    \end{subfigure}
    \hfill
    \begin{subfigure}[t]{0.48\linewidth}
       \includegraphics[width=\textwidth]{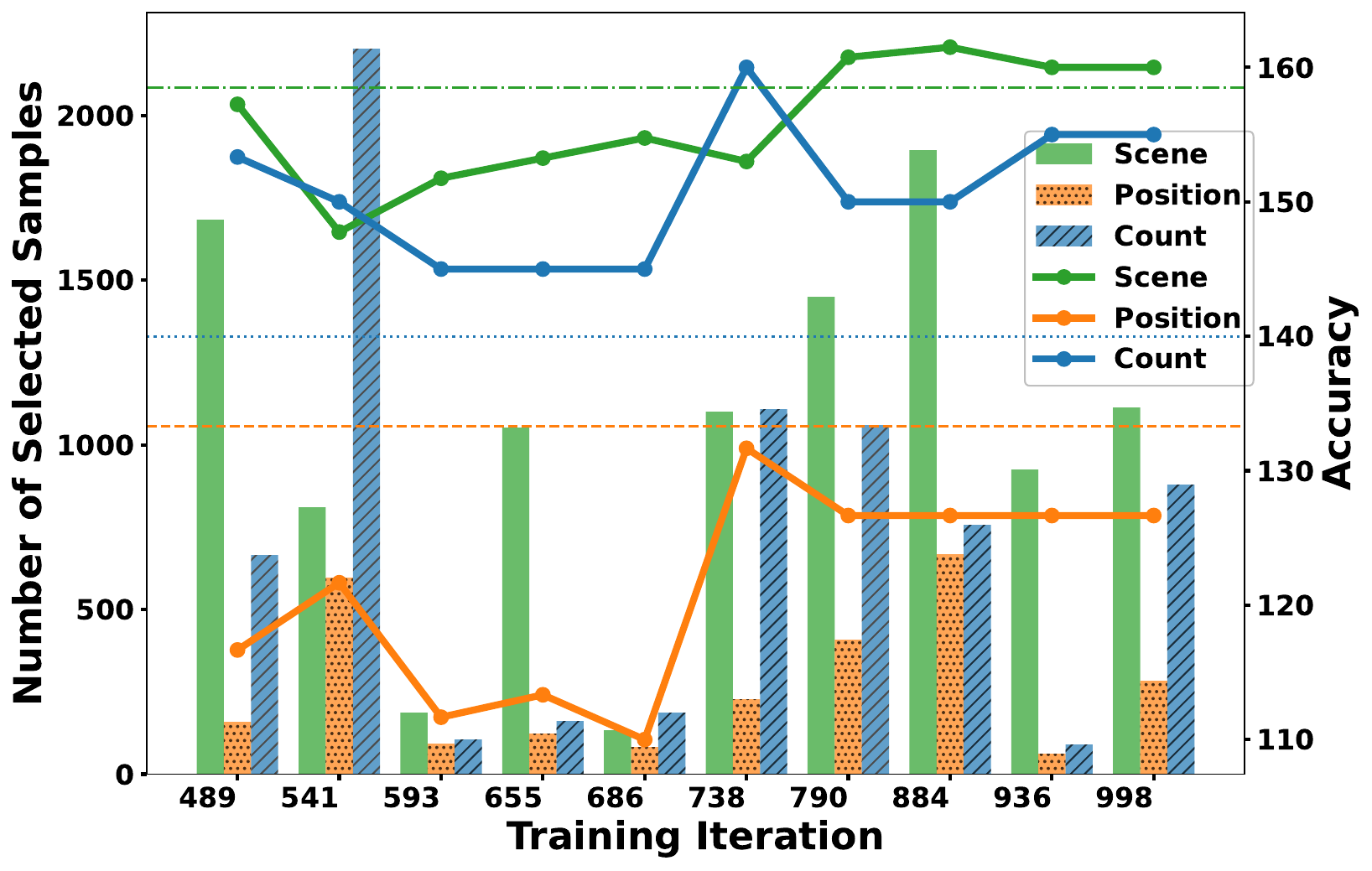}
       \caption{Perceptual Abilities}
       \label{fig:ability_easy} 
    \end{subfigure}
    \caption{\textbf{What skills model prioritize and when?}
    We track the number of samples selected per ability (bars) and corresponding accuracy trends during training, with dashed lines indicating full-data finetuning performance.
    }
    \label{fig:ability_learning}
    \vspace{-5mm}
\end{figure}

\noindent\textbf{What skills are prioritized during training?}
We analyze how different abilities emerge during training using fine-grained skill categories from prior work~\citep{liang2024survey}. Fig.~\ref{fig:ability_learning} groups clusters into \textit{language/symbolic} (OCR, translation, code) abilities and \textit{grounded perceptual} (scene, position, count). Each cluster is assigned a dominant ability (Appendix~\ref{app:what_and_when}), and we track both selected samples and accuracy over time.

For language abilities, OCR, a prerequisite for translation and code reasoning, is strengthened first: an increase in OCR-selected samples in mid-training (iterations 738–790) coincides with OCR recovering from an early dip and exceeding full-data performance. Subsequent budget allocation to \emph{translation} and \emph{code} leads to steady accuracy gains that remain above full-data baselines. 
For perceptual abilities, 
\emph{scene}—requiring coarse global recognition—acts as an easier prerequisite compared to \emph{count}, which also demands precise localization and enumeration.
The model initially prioritizes \textit{count} (at iteration 541), but accuracy does not improve — likely because it is too early in training to benefit from learning this skill. 
It temporarily shifts focus and improves other abilities (i.e \textit{scene}), then revisits \textit{count} and \textit{position} abilities at iteration 738, where accuracy now increases and later stabilizes.
Overall, this shows \ours consolidates easier, prerequisite abilities first and progressively invests its limited budget into harder skills.

% \vspace{-0.1in}
\section{Conclusion}
% \vspace{-0.1in}
We introduce \textsc{PROGRESS}, a dynamic and data-efficient framework for instruction-tuning VLMs under a strict data and supervision budget. By tracking learning progress across unsupervised skill clusters and prioritizing samples that are most learnable at each stage, \textsc{PROGRESS} effectively controls both the acquisition and order of skills. Our method achieves near full-data performance with just 16–20\% supervision while requiring no additional reference VLMs, requires annotations only on \emph{need basis} , and scales across architectures and datasets. Extensive experiments show that this self-paced, progress-driven strategy outperforms strong baselines while offering practical advantages in scalability, diversity, and data efficiency.

\clearpage
{
    \small
    \bibliographystyle{ieeenat_fullname}
    \bibliography{main}
}


% \vspace*{2em}
\clearpage
\newpage
{\LARGE \textbf{Appendix}}\\[1em]
In this appendix section, we provide additional details that could not be included in the main paper due to space constraints:
\begin{itemize}
\item Additional details on the \textbf{Baselines} (extending Sec.\ref{sec:exp_detail} and Fig. \ref{fig:comparison} of the main manuscript).
\item \textbf{Implementation} details and \textbf{hyperparameter} settings (extending Sec.\ref{sec:exp_detail}).
\item Further \textbf{Ablation} studies for \textbf{Hyperparameters} (extending Sec.\ref{sec:abl})
\item Analysis on \textbf{Why PROGRESS Works}: Representativeness
of Selected Subsets and Order of Skills
\item Comparison with concurrent work ICONS (extending Sec.\ref{results} and Sec \ref{related_work}).
{\item \textbf{Intuition and Justification} for using relative improvement (extending Sec.\ref{sec:method}).}
{\item In-depth analysis \textbf{Wall-clock Time, Time Breakdown Analysis, Annotation Time Analysis,} and  \textbf{Scaling Performance} (extending Sec.\ref{sec:abl})}.
\item \textbf{Word Cloud Visualization} for our multimodal clustering approach (extending Sec.\ref{sec:MM_cat}).
\item Visualization of the \textbf{Diversity} of samples selected by our method (extending Sec.~\ref{sec:abl}).
\item Additional details on the setup and algorithms used for our analysis (extending Sec.\ref{sec:analysis}).
\item Limitations of our method and LLM usage disclosure statement
\end{itemize}

\section{Details of Experimental Setups}
\label{app:exp}

\subsection{Baselines}
\label{app:baselines}
We follow the standard experimental settings and implementation protocols for all baselines as established in recent prior work (see COINCIDE \citep{coincide}), using official code. For completeness, clarity and reproducibility, we additionally provide detailed descriptions of each baseline here.

\textbf{COINCIDE} \citep{coincide} is a strong coreset selection method that leverages concept-based clustering and mutual transferability between clusters to guide sample selection. COINCIDE performs coreset selection only once, before training, by clustering internal activations of a separately trained (additional) reference VLM (e.g., TinyLLaVA\citep{zhou2024tinyllava}). COINCIDE relies on static selection strategy that does not adapt to the model's learning progress, requires an additional pretrained VLM, ground-truth annotations for full dataset to extract activation maps, and manual intervention to select appropriate activation layers—making the method resource-intensive and difficult to scale.

\textbf{CLIP-Score}~\citep{clipscore} It ranks image-instruction pairs based on visual-textual similarity computed by the CLIP model \citep{clip}, selecting top-scoring samples for training. While this approach assumes that higher similarity indicates greater informativeness, it relies on static, precomputed metrics that do not adapt to the model's learning progress. Prior work \citep{coincide} has shown that such metrics often fail to capture important data modes, resulting in reduced diversity and suboptimal generalization—limitations that \ours overcomes through dynamic, progress-driven selection.

\textbf{EL2N}~\citep{el2n} ranks training samples based on the expected L2-norm of prediction error:
\(
\mathbb{E}[ \| p(x) - y \|_2 ]
\)
where \( p(x) \) is the token distribution predicted by a reference VLM, \( x \) is the input, and \( y \) is the ground-truth label. This score reflects how confidently and accurately the reference model predicts each sample. However, it requires access to a fully trained additional VLM and ground-truth labels for the entire dataset, making it resource-intensive and static.

\textbf{Perplexity}~\citep{perplexity} measures the uncertainty in the model’s predictions and is defined as $\exp(-\mathbb{E}[\log p(x)])$, where $p(x)$ denotes the likelihood assigned to input $x$ by a additional reference VLM model. Samples from the middle of the perplexity distribution are selected, following prior work \citep{coincide}. However, it requires access to a fully trained additional VLM and, like other static metrics, often fails to capture important data modes—potentially limiting diversity and downstream generalization.

\textbf{SemDeDup}~\citep{semdedup} aims to reduce redundancy by removing semantically duplicated samples. It clusters the output embeddings of the final token from a reference model’s last layer and retains a diverse subset by eliminating near-duplicates and reducing redundancy. This method also requires an additional reference VLM to extract the final token features and ground-truth labels for the entire dataset.

\textbf{D2-Pruning} \citep{d2-prune} constructs a graph over training data where nodes encode sample difficulty and edges capture pairwise similarity. It selects a diverse coreset by pruning this graph while preserving representative and challenging samples. Difficulty is measured using the AUM score, defined as $p_y(x) - \max_{i \neq y} p_i(x)$, where $p_y(x)$ is the model’s confidence for the ground-truth label $y$. Similarity is computed using the L2 distance between average final-layer token embeddings from a additional reference VLM. This method requires access to an additional reference VLM for embedding extraction and scoring and ground-truth labels for the entire dataset.

\textbf{Self-Sup}~\citep{self-sup}  clusters data using averaged output embeddings from the final-layer tokens of a reference model. It assigns scores based on distance to cluster centroids, selecting samples that are closest—assumed to be the most prototypical representatives of the data distribution. This method also requires access to an additional reference VLM for embedding extraction.

\textbf{Self-Filter}~\citep{Chen2024selffilter} is a recent coreset selection method originally proposed for the LLaVA-158k dataset (containing three vision-language tasks). It jointly fine-tunes a scoring network alongside the target VLM on the \emph{entire labeled dataset}, using this as learned reference model to score and filter training samples—hence it requires an additional reference model trained on full data with full annotations. Following previous work \citep{coincide}, we adopt the stronger variant that also incorporates CLIP scores and features.

\textbf{Random.} We additionally report results for \textit{Random}, which finetunes the model using a coreset selected via random sampling. Despite its simplicity, Random serves as a strong and competitive baseline—prior work \citep{coincide}has shown that random sampling often preserves sample diversity and can outperform more complex selection methods in certain settings.

\textit{Note:} We use standard setup for the baseline implementations as described in prior work
(see COINCIDE Appendix \citep{coincide}). For \textbf{COINCIDE}, \textbf{EL2N}, \textbf{SemDeDup}, \textbf{D2-Pruning}, and \textbf{Self-Sup}, we use image, question, and ground-truth answer for full dataset as inputs along with additional reference VLM (i.e., TinyLLaVA) to extract features following prior work \citep{coincide}. 
\textbf{Self-Filter} requires full dataset to finetune additional reference network—the score-net.
As a result, these baselines require an additional reference vision-language model or full dataset annotations (100\%) or both.

\vspace{0.1in}
\noindent\textbf{Active Learning Nature of the Compared Baselines.}
% \paragraph{Active Learning Baselines}
AL selects which unlabeled samples to label next using acquisition function a(x) \citep{weng2022learning}. Higher a(x) (e.g., uncertainty, error, diversity, representativeness) implies higher expected utility of sample.
EL2N \citep{el2n} and Perplexity \citep{perplexity} are uncertainty-based AL methods with error/uncertanity as acquisition functions.
Self-Filter \citep{Chen2024selffilter} uses deep network as acquisition function. It jointly trains a deep score network with the target model to later rank and select samples.
COINCIDE \citep{coincide} is hybrid sampling based AL which uses a hybrid of skill transferability + diversity for acquisition of samples to select a representative coreset.
Self-Sup \citep{self-sup} aligns with representativeness-based AL coreset selection, aiming to pick prototypical samples that best cover the data distribution.
% ICONS uses gradient-based scores as acquisition function.

\vspace{0.1in}
{\noindent\textbf{Curriculum Learning Nature of Compared Baselines.}}
Curriculum Learning (CL) methods can be partitioned into two types:
(Type 1) External model decides difficulty and select samples; 
(Type 2) Models own feedback decides difficulty and select samples.

Self-Filter \citep{Chen2024selffilter} is an instantiation of Type 1 CL methods where an external model (i.e deep score network) is jointly trained with the target model to quantify the learning difficulty of sample and select hard-examples for training. The auxiliary score network guides selection of samples expected to have significant value for model training based on difficulty.

We also compare with Type 2 curriculum policies in Table \ref{tab:ablation_policy}:
1) Easiest Selection (Table \ref{tab:ablation_policy} row 3) - model self-evaluates and chooses easy-samples with highest absolute performance,
2) Medium difficulty Selection (Table \ref{tab:ablation_policy}, row 4)- model self-evaluates and chooses samples with medium absolute performance,
3) Hardest Selection (Table \ref{tab:ablation_policy}, row 5) - model self-evaluates and chooses hard-samples with lowest absolute performance
4) Relative-Improvement based selection (ours, Table \ref{tab:ablation_policy}, row 6-7). \\
Overall, our PROGRESS method (relative improvement based policy) performs much better than other CL baselines.

\subsection{Implementation Details}
\label{app:impl}

\begin{table}
\centering
\caption{Hyperparameter configurations. $K$ represents the number of clusters.}
% \resizebox{0.5\textwidth}{!}{  % slight margin to fit within 0.5\textwidth
    \renewcommand{\arraystretch}{1.25}
    \renewcommand{\tabcolsep}{3pt}
     \adjustbox{max width=\columnwidth}{
    \begin{tabular}{lll}
    \toprule
    Method  & LLaVA-1.5-665K & Vision-Flan \\ 
    \midrule
    CLIP-Score & high score selected & high score selected \\
    EL2N & medium score selected & medium score selected \\
    Perplexity & medium score selected & medium score selected \\
    SemDeDup & $\bm{K}$\;:\;10,000 & $\bm{K}$\;:\;5,000 \\
    D2-Pruning & $\bm{k}$\;:\;5,\;$\gamma_r$\;:\;0.4,\;$\gamma_f$\;:\;1.0 & $\bm{k}$\;:\;5,\;$\gamma_r$\;:\;0.4,\;$\gamma_f$\;:\;1.0 \\
    Self-Sup & $\bm{K}$\;:\;10,000 & $\bm{K}$\;:\;5,000 \\
    Self-Filter & $\bm{k}$\;:\;10,\;$\gamma$\;:\;1 & $\bm{k}$\;:\;10,\;$\gamma$\;:\;1 \\
    COINCIDE & $\bm{K}$\;:\;10,000,\;$\tau$\;:\;0.1 & $\bm{K}$\;:\;5,000,\;$\tau$\;:\;0.1 \\
    \midrule
    \textbf{\ours} & \\
    \textbf{Warmup Stage }   \\
    Number of Clusters  & $\bm{K}$\;:\;10,000  & $\bm{K}$\;:\;5,000  \\
    Warmup Ratio (w.r.t full data) & 9\%   &  8.4\%  \\
    \textbf{Prioritized Concept Learning}  \\
    Number of Clusters &$\bm{K}$\;:\;1,000  & $\bm{K}$\;:\;200\\
    Temperature of Softmax & $\tau$\;:\;1.0 & $\tau$ \;:\;1.0\\
    BatchSize &128  & 128\\
    
    Selection Gap & $\bm{\gamma}*BatchSize$\;:\;7,500  & $\bm{\gamma}*BatchSize$\;:\;3,500\\
     Random Exploration & \( \delta\% \)\;:\; 10 \%  & \( \delta\% \)\;:\; 10 \%\\
    \bottomrule
    \end{tabular}
   }
% }
\vspace{-0.1in}
\label{tab:hyper}
\end{table}

In this section, we provide elaborate details on implementation of our approach in continuation to the brief details we provide in Section \ref{sec:exp_detail} of main manuscript.

We first partition the unlabeled data pool \( \mathbb{U} \) into \( K \) skill clusters using spherical k-means, following the fully \textit{unsupervised} concept categorization procedure described in Section~\ref{sec:MM_cat}. Training begins with a brief warmup phase (see details in Appendix~\ref{sec:warmup}), which equips the model with basic instruction-following capability and ensures that skill-level performance estimates are reliable in the beginning of training.

Subsequently, we apply our Prioritized Concept Learning (PCL) strategy (see Section~\ref{PCL}) to estimate the expected performance improvement for each skill cluster between iteration \( t \) and \( t - \gamma \) (as defined in Eqn~\ref{eq:delta}), using either accuracy or loss as the tracking metric (see Table~\ref{tab:llava_eval}, row 10 and 11). For the accuracy-based variant, we compute cluster-wise accuracy using an LLM judge as metric that compares the VLM output to ground-truth answers—though this is not required for our loss-based variant. Samples are then selected using a temperature-controlled softmax over the improvement scores (see Eqn~\ref{eq:softmax}). This selection process is repeated every \( \gamma \) iterations, and in each round, we sample a total of \( \gamma * {BatchSize} \) examples for annotation and training. We refer to this \( \gamma * {BatchSize} \) as selection gap from here on.

\paragraph{Hyperparameters for Baselines and \ours.}
To ensure fair comparison, we use the same hyperparameters as COINCIDE~\citep{coincide} for all baselines. The hyperparameters for both the baselines and \ours are summarized in Table \ref{tab:hyper}.
For model training, we apply LoRA~\citep{hu2021lora} to LLaVA-v1.5 and follow the official fine-tuning settings provided in the LLaVA-1.5 release. For Qwen2-VL, we perform full fine-tuning using the official hyperparameters specified by Qwen2-VL.
{For Qwen2.5-32B-Instruct, we apply LoRA and follow the official fine-tuning settings provided in LLaMA-Factory \citep{zheng2024llamafactory}.}
For accuracy estimation, we use LLMs such as InternLM2-Chat-20B~\citep{cai2024internlm2} as the judge. We provide the question, ground-truth answer, and predicted answer (without the image) as input and ask the LLM to decide whether the prediction is correct. The full prompt is shown below.
Ablation studies on all hyperparameters are provided in Section~\ref{sec:abl} and Appendix~\ref{app:ana}.

\begin{figure}[t]
\centering
\begin{tcolorbox}[colback=gray!5!white, colframe=gray!75!black, title=Prompt for Accuracy Estimation]
Given an input question and two answers: a candidate answer and a reference answer, determine if the candidate answer is correct or incorrect.

\textbf{Rules:}
\begin{itemize}
    \item The candidate answer is correct if it is semantically equivalent to the reference answer, even if they are phrased differently.
    \item The candidate answer should be marked as incorrect if it:
    \begin{itemize}
        \item Contains factual errors compared to the reference answer
        \item Only partially answers the question
        \item Includes hedging language (e.g., "probably", "likely", "I think", etc.)
        \item Answers a different question than what was asked
    \end{itemize}
    \item Give a reason for your prediction.
\end{itemize}

\textbf{Output Format:}
\begin{itemize}
    \item Answer - correct or incorrect
    \item Reason -
\end{itemize}
\label{tab:prompt}
\end{tcolorbox}
\end{figure}

\subsection{Warmup Phase}
\label{sec:warmup}

\paragraph{Warmup strategy details.}
Following prior work~\citep{xialess,icons}, we begin with a brief warmup phase using a small subset—9\% of the total pool size (we show in Fig.~\ref{fig:ablation}(a) that \ours is robust to even smaller warmup ratios, remaining close to full-data performance even when the warmup percentage is further reduced.)
— to equip the model with basic instruction-following capability and to obtain credible skill-level performance estimates at the start of training, when the model is still untrained. These initial estimates are essential for tracking relative learning progress across skills in subsequent phases. To make the warmup set effective, it should broadly cover diverse skill clusters. 

To this end, we use a simple cluster-based sampler that prioritizes clusters which are both diverse and likely to generalize well. Concretely, we sample from each unsupervised concept cluster with probability
$
P_i \propto \exp\left({S_i}/{\tau D_i}\right)$
where $S_i$ and $D_i$ denote the cluster’s transferability and density, respectively, and $\tau$ is temperature following~\cite{coincide}. Importantly, unlike ~\cite{coincide}, our clusters and scores are computed directly from concatenated self-supervised DINO and BERT features (Sec.~\ref{sec:MM_cat}), and do \emph{not} require any additional auxiliary reference VLM (i.e TinyLLaVA) features or ground-truth answers. 

Table~\ref{tab:ablation_warmup} (main manuscript) shows that the warmup phase alone provides only modest performance (94.6\% Rel.), while \ours yields large gains over this baseline (+4\% relative, rows 2 vs.\ 1). This demonstrates that warmup merely initializes reasonable per-cluster estimates, whereas the major improvements stem from \ours's progress-driven sample selection. Moreover, row 3 shows that \ours remains robust to different warmup strategies, achieving near–full-data performance even with a weak (random sampling) warmup.

\begin{table*}[t]
    \tiny
    \caption{Statistics of ICONS target validation sets.
    }
    \centering
\resizebox{0.8\textwidth}{!}{
\begin{tabular}{l|ccccccccccc}
\toprule
\textbf{Dataset} & MME & POPE & SQA-I & MMB-en & MMB-cn & VQAv2 & GQA & VizWiz & TextVQA & LLaVA-W \\
\midrule
$|\mathcal{D}_{\text{val}}|$ & 986 & 500 & 424 & 1,164 & 1,164 & 1,000 & 398 & 8,000 & 84 & 84 \\
$|\mathcal{D}_{\text{test}}|$ & 2,374 & 8,910 & 4,241 & 1,784 & 1,784 & 36,807 & 12,578 & 8,000 & 5,000 & 84 \\
\bottomrule
\end{tabular}
}
\label{tab:icons_target}
\end{table*}

\begin{table*}[t]
    \tiny
    \caption{Comparison between \ours and ICONS. Repro. means reproductions of ICONS.
    }
    \centering
    {
        \renewcommand{\arraystretch}{1.3}
        \renewcommand{\tabcolsep}{3.0pt}
        \begin{tabular}{l c c c c c c c cc c c c c c c}
             \toprule
             {\textbf{Method}} & {\textbf{VQAv2}} & {\textbf{GQA}} & {\textbf{VizWiz}} & {\textbf{SQA-I}} & {\textbf{TextVQA}} & {\textbf{POPE}} & {\textbf{MME}} & \multicolumn{2}{c}{\textbf{MMBench}} & {\textbf{LLaVA-}}  &{\textbf{SEED}}  & {\textbf{AI2D}}   & {\textbf{ChartQA}} & {\textbf{CMMMU}}  &  \cellcolor{gray!25}{\textbf{Rel. (\%)}}\\
             & & & & & & & & {\textbf{en}} & {\textbf{cn}} & {\textbf{Wild}} &  & \\
             \midrule
             \ 0\ \ \ Full-Finetune &
             {\scriptsize 79.1} & {\scriptsize 63.0} & {\scriptsize 47.8} & {\scriptsize 68.4} & {\scriptsize 58.2} & {\scriptsize 86.4} & {\scriptsize 1476.9} & {\scriptsize 66.1} & {\scriptsize 58.9} & {\scriptsize 67.9}  & {\scriptsize 67.0} & {\scriptsize 56.4} & {\scriptsize 16.4} & {\scriptsize 22.1} & \cellcolor{gray!25}{\scriptsize 100}\\
             \cmidrule{0-15}
             \cmidrule{0-15}
             \ 1\ \ \ ICONS & {\scriptsize 76.3} & {\scriptsize 60.7} & {\scriptsize 50.1} & {\scriptsize 70.8} & {\scriptsize 55.6} & {\scriptsize 87.5} & {\scriptsize 1485.7} & {\scriptsize 63.1} & {\scriptsize 55.8} & {\scriptsize 66.1} & {\scriptsize -} & {\scriptsize -} & {\scriptsize -} & {\scriptsize -} & \cellcolor{gray!25}{\scriptsize -}\\
             \ 2\ \ \ ICONS (Repro.) & {\scriptsize 75.0} & {\scriptsize 57.7} & {\scriptsize 45.9} & {\scriptsize 63.7} & {\scriptsize 55.1} & {\scriptsize 86.0} & {\scriptsize 1434.0} & {\scriptsize 47.1} & {\scriptsize 37.3} & {\scriptsize 68.4} & {\scriptsize 57.3} & {\scriptsize 45.3 } & {\scriptsize 17.2} & {\scriptsize 24.3 } & \cellcolor{gray!25}{\scriptsize 91.6 }\\
             \textbf{\ours } \\ 
             \ 3\ \ \ Loss as Obj. & {\scriptsize 75.7} & {\scriptsize 58.6} & {\scriptsize 49.6} & {\scriptsize 70.1} & {\scriptsize 55.1} & {\scriptsize 86.3} & {\scriptsize 1498.4} & {\scriptsize 62.5} & {\scriptsize 55.5} & {\scriptsize 65.5} & {\scriptsize 63.4} & {\scriptsize 53.3} & {\scriptsize 17.3} & {\scriptsize 23.7} & \cellcolor{gray!25}{\scriptsize {98.4}} \\
             \ 4\ \ \  Accuracy as Obj. & {\scriptsize 75.2} & {\scriptsize 58.8} & {\scriptsize 53.4} & {\scriptsize 69.9} & {\scriptsize 55.1} & {\scriptsize 85.9} & {\scriptsize 1483.2} & {\scriptsize 61.1} & {\scriptsize 54.4} & {\scriptsize 65.5} & {\scriptsize 63.0} & {\scriptsize 52.8} & {\scriptsize 17.3}  & {\scriptsize 24.6} & \cellcolor{gray!25}{\scriptsize {98.8}} \\
             \bottomrule
  
        \end{tabular}
    }
    \label{tab:icons_eval}
    \vspace{-5mm}
\end{table*}

\subsection{Comparison with ICONS}
\label{app:icons}

\textbf{ICONS}~\citep{icons} is a concurrent unpublished work that differs significantly from our approach. It requires \textbf{(1)} high memory and compute resources—reportedly over 100 GPU hours—to compute and store gradient-based influence scores for selection, and \textbf{(2)} access to explicit knowledge of the target task or its distribution in the form of labeled samples from validation set of target benchmarks. This assumption is impractical in general-purpose VLM training, where target tasks may be unknown at training time and usage of such high-compute refutes the goal of efficient learning. As such, ICONS is \textbf{not directly comparable} and falls outside the scope of our setting, which avoids both gradient-based selection and downstream task knowledge from target benchmarks prior to training the VLM model.

Nevertheless, we strive to compare with them in good faith by reproducing ICONS using their official codebase for fair comparison. Although ICONS has released its codebase, the validation data (for each target benchmark) it uses to simulate target task knowledge is not publicly available. The paper reports the number of validation samples used per benchmark (see Table~\ref{tab:icons_target}), however the specific validation samples remain unspecified and are not publicly released. To approximate their setup, we randomly select an equal number of samples from the publicly available validation sets of target benchmarks and reproduce their performance.

Table~\ref{tab:icons_eval} presents results across three settings: Row 1 shows the original ICONS results as reported; Row 2 presents our reproduction using their codebase and randomly selected validation samples; Rows 3 and 4 report results for \ours. We observe that \ours outperforms our ICONS reproduction in relative performance even though our method does not rely on compute-intensive gradient-based selection and does not assume any knowledge from target benchmarks, reinforcing the practicality and effectiveness of our method under realistic constraints.

\subsection{Intuition, justification, and grounding for using relative improvement and softmax-based sampling}
\label{app:rel_and_softmax}

\textbf{Empirical Validation.} We show ablation results and discussion in Table \ref{tab:ablation_policy} for different selection policy including Our Relative improvement strategy (Table \ref{tab:ablation_policy}, row 6,7) vs Easy, Medium and Hard selection based on hard-thresholding on absolute performance. Overall, our relative improvement policy performs much better than other variants

\textbf{Intuition, justification for using relative improvement.} Inspired by prior curriculum literature \citep{lba,sachan2016easy,bengio2009curriculum} we use relative improvement over absolute gains because it scales progress by a skill’s baseline performance i.e normalizes across tasks of different difficulty and scale: tasks that are too easy (high baseline) or too hard (low baseline) tend to have small relative improvements, whereas tasks at a “moderate” difficulty yield the largest relative improvements and therefore get prioritized.
This aligns with the zone‑of‑proximal‑development \citep{vygotsky2012thought,DataEnvGym}, problems that are neither too easy nor too hard produce the largest advantage, focusing on learning where progress is most productive.

In contrast, using absolute improvement would bias selection towards tasks with high raw score over‑favoring raw score jumps and fail to account for the fact that a one‑point increase on a hard skill is more meaningful than the same gain on an easy one, fixed thresholds likewise cannot adapt to varying skill difficulties. Overall, relative improvement avoids over‑favoring weak skills with noisy raw swings and instead emphasizes areas where progress per unit of remaining error is highest, providing a stable, self‑paced curriculum \citep{lba,sachan2016easy}.

\textbf{Intuition for Softmax Sampling.} We map $\Delta_k^t$ to sampling weights using a temperature-controlled softmax, $p_k \propto \exp(\Delta_k^t / \tau)$, which provides a smooth, entropy-regularized trade-off between exploiting high-utility skills and exploring others. A small floor probability safeguards rare skills. This results in a stable, noise-robust scheduler that outperforms absolute gains and hard thresholds. The temperature parameter $\tau$ explicitly balances informativeness and diversity which is crucial for effective learning and avoid mode collapse (see Fig \ref{fig:abl_temp}).

\section{Further Ablation Studies and Analysis}
\label{app:ana}

\subsection{Scalability and Generalization to Larger VLMs}
\label{app:qwen_32b}

% \paragraph{}

{Here we additionally present results for Qwen2.5-VL-32B-Instruct in Table~\ref{tab:qwen_25_32b}, which we instruction-tuned using our method on the LLaVA-665K dataset, using a 20\% sampling ratio following standard protocol. \ours (trained with just 20\% data) performs better than full- data finetuning, achieving  100.2\% relative performance compared to full-data fine-tuning, demonstrating our method's scalability potential and its generalization to new large scale architectures.}
\begin{table*}[t]
    \tiny
    \caption{\textbf{Scalability and Fine-tuning.} 
    We report results for Qwen2.5-VL-32B on the LLaVA-665K dataset using 20\% sampling ratio.
    }
    \centering
    {
        \renewcommand{\arraystretch}{1.25}
        \renewcommand{\tabcolsep}{5.0pt}
        \begin{tabular}{l c c c c c c c cc c c c}
             \toprule
             {\textbf{Method}}  & {\textbf{VQAv2}}  & {\textbf{GQA}} & {\textbf{VizWiz}} & {\textbf{SQA-I}} & {\textbf{TextVQA}} & {\textbf{POPE}} & {\textbf{MME}} & \multicolumn{2}{c}{\textbf{MMBench}} & {\textbf{LLaVA-}} & {\textbf{SEED}} & \cellcolor{gray!25} {\textbf{Rel. (\%)}}\\
             & & & & & & & & {\textbf{en}} & {\textbf{cn}} & {\textbf{Wild}} & \\
             % \midrule
             % \multicolumn{11}{c}{\textbf{Architecture Generalization  (Qwen2-VL-7B)}} \\
             \midrule
             Full-Finetune & {\scriptsize 83.8} & {\scriptsize 64.5} & {\scriptsize 58.8} & {\scriptsize 93.1} & {\scriptsize 82.3} & {\scriptsize 88.5} & {\scriptsize 1678.5} & {\scriptsize 85.3} & {\scriptsize 85.3} & {\scriptsize 78.9} & {\scriptsize 75.8} & \cellcolor{gray!25}{\scriptsize 100.0}\\

             \textbf{\ours}   & {\scriptsize 83.7}   & {\scriptsize 63.7} & {\scriptsize 59.0} & {\scriptsize 93.1} & {\scriptsize 82.0} & {\scriptsize 88.1} & {\scriptsize 1684.8} & {\scriptsize 86.7} & {\scriptsize 87.1} & {\scriptsize 77.8} & {\scriptsize 76.4}  &\cellcolor{gray!25} {\scriptsize \textbf{100.2}}\\
             \bottomrule
  
        \end{tabular}
    }
    \label{tab:qwen_25_32b}
\end{table*}

\subsection{Ablation Studies}
\label{app:ablation}

\begin{figure*}[t]
    \centering
    \begin{subfigure}[b]{0.26\linewidth}
        \includegraphics[width=\linewidth]{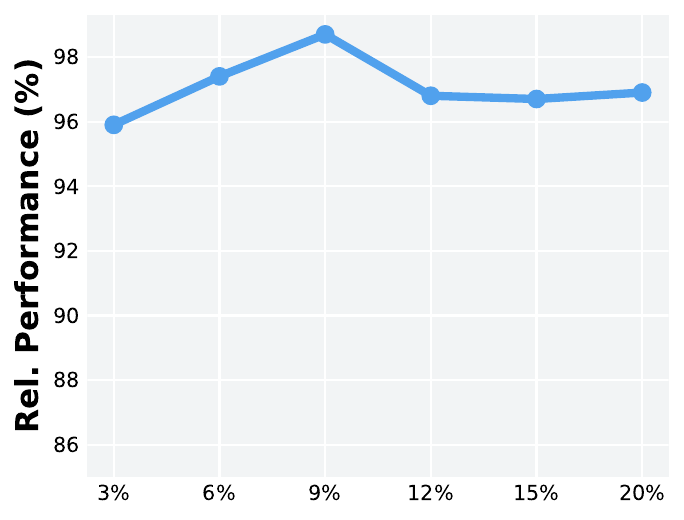}
        \caption{Warmup Ratio}
        \label{fig:abl_warmup}
    \end{subfigure}
    \begin{subfigure}[b]{0.26\linewidth}
        \includegraphics[width=\linewidth]{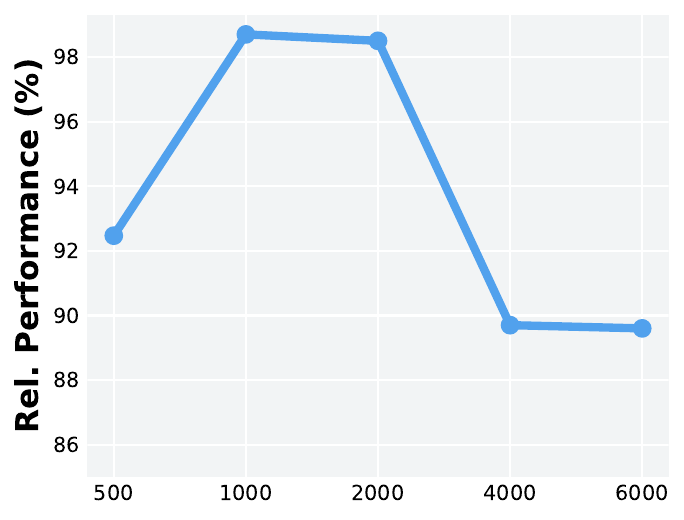}
        \caption{Cluster Number $K$}
        \label{fig:abl_cluster_num} 
    \end{subfigure}
    \begin{subfigure}[b]{0.26\linewidth}
        \includegraphics[width=\linewidth]{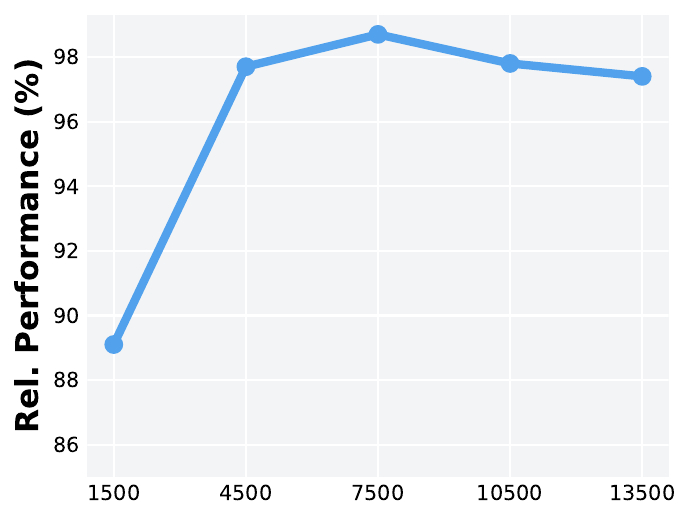}
        \caption{Selection Gap \(\gamma\)*$BatchSize$}
        \label{fig:abl_selection_gap}
    \end{subfigure}
    \caption{\textbf{Ablation Studies.}
(a) Effect of the warm-up ratio.
(b) Effect of the number of clusters.
(c) Effect of the selection gap.}
    \label{fig:ablation}
    % \vspace{-3mm}
\end{figure*}

\paragraph{Ablations on Hyperparameter.}
We conduct further ablations studies in Fig. \ref{fig:ablation} on effect of hyperparameters.
All experiments use LLaVA-v1.5-7B on the LLaVA-665K dataset with 20\% sampling ratio and accuracy as the objective.
Fig. \ref{fig:ablation}(a) shows the effect of different warm-up ratios relative to the total training data pool size. Our results show that a 9\% warm-up ratio achieves the best performance, as it strikes a balance between preparing the model adequately and leaving enough room for our iterative Prioritized concept learning strategy to select informative samples. The 20\% warm-up ratio (i.e using only warmup selected samples to entirely train the model eleminating our Prioritized concept learning strategy completely), results in significantly reduced performance in overall relative score.
Next, Fig. \ref{fig:ablation}(b) shows the effect of varying the number of clusters $K$ used for our concept categorization module (Section~\ref{sec:MM_cat}). Using too few skill-clusters reduces skill diversity and leads to lower purity (in terms of skill types) within  given cluster, while too many clusters result in redundant clusters of the same skill category and insufficient samples per cluster to yield credible accuracy estimates. We find that using approximately 1,000-2000 clusters strikes the best balance and yields optimal performance.
Finally, Fig. \ref{fig:ablation}(c) shows the influence of the selection gap i.e \(\gamma\) *$BatchSize$ (see definition in Appendix \ref{app:impl}). 
We find that the model is particularly sensitive to small gaps; for instance, a gap size of 1,500 leads to a rapid performance decrease. Smaller gaps cause the model to switch too soon, not allowing it to learn the selected concepts sufficiently.

\noindent\textbf{\large Why \ours Works: Representativeness of Selected Subsets and Order of Skills}

\noindent\textbf{Representativeness of the Selected Subset.}
We evaluate how well the subsets selected by \ours and COINCIDE reflect the full dataset using PCA-based analyses. We first concatenate the DINO and BERT features and fit a single Principal Component Analysis (PCA) model on the entire dataset, reducing the dimensionality to 50 components. This global PCA transform is then applied consistently to the full dataset and to the samples selected by both methods. In this 50-dimensional PCA space, we compute two measures of representativeness:
\begin{itemize}
    \item \textbf{Mean Nearest-Neighbor (NN) Distance.}
To estimate how densely each subset covers the data manifold, we build a FAISS \texttt{IndexFlatL2} index containing all PCA-transformed points from the full dataset. For every point in a subset, we query this index to find its closest neighbor \emph{within the full dataset} and measure that Euclidean distance. Averaging these distances yields a single value that reflects how close the subset stays to the dense, frequently occurring regions of the distribution. 
Lower values indicate that the subset lies in typical, well-populated regions rather than isolated or sparsely sampled areas.
\ours achieves a mean NN distance of $|$0.083$|$, substantially lower than COINCIDE’s $|$0.198$|$, indicating that \ours selects samples closer to the core of the dataset.\\
While mean NN distance is intuitive and meaningful, it can be minimized by degenerate subsets that concentrate in a few high-density regions and fail to preserve the global structure of the distribution. For this reason, we also report a complementary \emph{Variance Coverage} metric.

\item \textbf{Variance Coverage.}  
PCA orders directions in the feature space by how much variation they capture: the leading components encode the dominant structural patterns of the dataset, while later components reflect finer-grained variations. To assess how well each subset preserves this structure, we compare its variance along every principal component to the corresponding variance in the full dataset. For each component, we compute  
\(
\text{ratio} = \frac{\text{variance (subset)}}{\text{variance (full)}},
\)
which measures how much of the dataset’s variation along that axis the subset retains.  
We summarize this comparison in two ways. First, we compute a \emph{Weighted Variance Retained} score by weighting each variance ratio by the component’s explained variance in the full dataset and summing across all 50 components. This yields a single measure of how much of the dataset’s overall variability the subset preserves. \ours retains more variance than COINCIDE (38.7\% vs.\ 34.2\%).  
Second, we directly examine the variance ratios for the top principal components, which capture the major structural axes of the dataset. High coverage on these components indicates that a subset spans the most important modes of variation rather than collapsing onto a narrower region. \ours covers a substantially larger fraction of these leading components compared with COINCIDE (77–88\% vs.\ 48–72\%).

\end{itemize}

These results show that \ours spans the overall data distribution more effectively, while COINCIDE selects a more concentrated and less representative subset.

\noindent\textbf{How important is the order of skill acquisition?}
Unlike prior methods that focus solely on selecting which samples to use \citep{coincide,icons}, \ours also controls when to introduce them during training (Section \ref{PCL})—effectively guiding both skill selection and the order of acquisition.
To assess the importance of learning order, we ablate this component by randomly shuffling the data selected by \ours and training the model without respecting the intended sequence.
Even with the same data, training in a random order leads to a noticeable performance drop—from 98.8\% to 94.6\%—highlighting that when to introduce concepts is just as important as what to learn.

\begin{wrapfigure}{r}{0.25\textwidth}
    \vspace{-8mm}
    \begin{center}
    \includegraphics[width=0.25\textwidth]{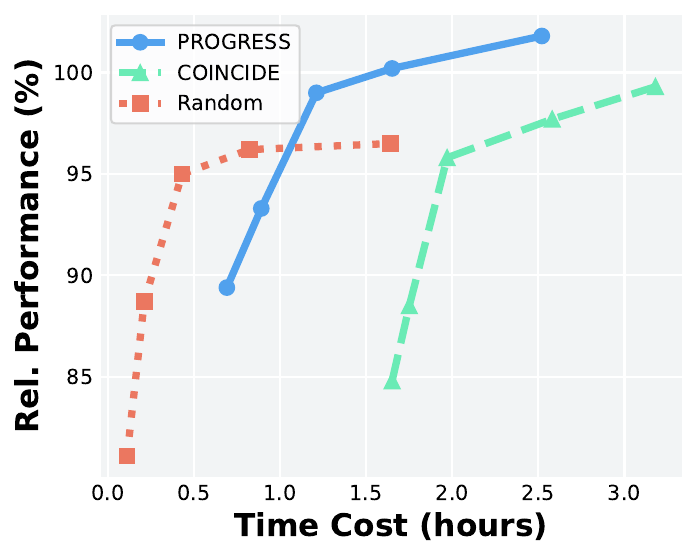}
    \caption{\textbf{Wall-clock time} comparison on Vision-Flan dataset.}
    \label{fig:wall_clock_time_app}
    \end{center}
    \vspace{-3mm}
\end{wrapfigure}

\subsection{Wall-clock Time Comparison}
\label{app:wall_clock_app}
We compare average relative performance (Rel.) against wall-clock time on the Vision-Flan dataset in Fig. \ref{fig:wall_clock_time_app}.
\ours (Accuracy as Objective) achieves relative performances of 89.4\%, 93.3\%, 99.0\%, 100.2\% and 101.8\% with wall-clock times of 0.69, 0.89, 1.21, 1.65 and 2.52 hours.
Full fine-tuning on entire dataset takes about 2.56 hours.
Remarkably, our method exceeds 100\% relative performance (i.e surpasses vanilla full data fine-tuning) in just 1.65 hours—including both data selection and model training time—needing only 64\% of the time required for full dataset fine-tuning.

\begin{table*}[h]
\centering
\caption{\textbf{Time breakdown comparison:} COINCIDE vs PROGRESS vs Full Data Training. We use official code to produce this time analysis}
\label{tab:time_comparison}
\adjustbox{max width=0.7\textwidth}{
\begin{tabular}{p{3.7cm}p{2.5cm}p{6.5cm}}
\toprule
\textbf{Stage} & \textbf{Time} & \textbf{Note} \\
\midrule
\textbf{\textit{Full Data Training}} & \textbf{540 min (9h)} & Training on 100\% data on 4×A100 \\
\midrule
\multicolumn{3}{l}{\textbf{\textit{COINCIDE}}} \\
\midrule
Manual Inspection & Unknown & Manual MSA layer selection to extract features, needs ground-truth annotations\\
Feature Extraction & 280 mins & TinyLLaVA feature extraction on 2×A100 \\
Clustering & 50 mins &  	With faiss-gpu KNN on single A100. COINCIDE requires large memory to store 20,480 dim. features from multiple MSA layers. \\
Training & 150 mins & Training on 20\% data on 4×A100 \\
% \midrule
\textbf{Total} & \textbf{480 min (8h)} & \\
\midrule
\multicolumn{3}{l}{\textbf{\textit{PROGRESS}}} \\
\midrule
Feature Extraction & 30 mins & Sentence-BERT (all-MiniLM-L6-v2) + DINO-v2-base on single A100 \\
Clustering & 30 mins & With faiss-gpu KNN on single A100 \\
Annotation Decision & 130 mins & Model self-evaluates and decides what to annotate next. 13 mins × 10 rounds  \\
% \midrule
Training & 150 mins & Training on 20\% data on 4×A100 \\
\textbf{Total} & \textbf{340 min (5.67h)} & \\
% \midrule
% \multicolumn{3}{l}{\textbf{\textit{Baseline}}} \\
\bottomrule
\end{tabular}
}
\end{table*}

\subsection{Stage-Wise Time Breakdown Analysis}
\label{app:time_break_down}

{Table \ref{tab:time_comparison} presents a detailed time breakdown comparing our PROGRESS method with the closest competitor COINCIDE. 
We use same GPU compute for both COINCIDE and our method for fair comparison.
COINCIDE requires over 8 hours of total computation time (plus additional unknown manual inspection time) and needs an additional pre-trained TinyLLaVA VLM to extract and store features from multiple MSA layers (4th, 8th, 12th, 16th, 20th). For each sample, they need 4096×5 = 20,480 dimensional features, which demand substantial memory resources and significantly increases clustering time. In contrast, our PROGRESS method completes in only 5.67 hours without requiring any additional VLM models. By using lightweight uni-modal self-supervised feature extractors—combining DINO-v2 (1,024 dimensions) and Sentence-BERT (384 dimensions)—we achieve efficient clustering with only 1,408-dimensional features per sample. This represents a 14.5× reduction in feature dimensionality compared to COINCIDE, while still achieving better relative performance w.r.t COINCIDE using only 20\% of the training data. }

\subsection{Annotation Time Analysis: Significance of Annotation Cost as the Primary Bottleneck}
\label{app:annotation_time}
In this section, we estimate the overall annotation cost and show that it is the dominant bottleneck in scaling VLM training to larger datasets—a cost that our method drastically reduces by 80\% (as it requirs only 20\% data for training).

The LLaVA-665K dataset has 0.67 million samples comprising Human-curated data (OKVQA, A-OKVQA, OCRVQA, TextCaps)  and  Synthetic QA which still need human-annotated bounding box \& object names (from COCO etc, see details in \citep{llava,liu2024llavanext}) which are then provided to an LLM to generate Q/A.

Prior studies \citep{blog2020simple} estimate it typically takes 10.3 sec for a human (Mturk worker) to annotated 1 sample (which consists Image \& 4-5 Q/A pairs). So, to annotate 0.67 million samples in LLaVA dataset it will take 1902 hours (0.67 M * 10.3 / 3600).

\textbf{Estimated annotation time:}  
\hspace{0.5em}\\• Full-data training (100\% of 0.67M samples): $\sim$1902 hr  
\hspace{0.5em}\\• \ours (20\% of 0.67M samples): $\sim$380 hr  

\textbf{Overall time (annotation + training):}  
\hspace{0.5em}\\• Full-data finetuning: 1902 hr (annotation) + 9 hr (training)  
\hspace{0.5em}\\• \ours: 380 hr (annotation) + 5.7 hr (training)

\textbf{Conclusion.} Annotation is by far the dominant cost and major bottleneck in VLM training, especially when we scale to even larger datasets. By reducing annotation to only 20\%, \ours drastically cuts the primary bottleneck and offers a scalable path for training on even larger datasets.

\subsection{Scaling data improves PROGRESS even more than vanilla Full-Finetuning}
\label{app:scaling_data}

We point to results shown in Table \ref{tab:abl_scaling} and Fig \ref{fig:sample_ratio} where scaling data size to higher percentages 32\%, 64\%, our method outperforms full-finetuning (which uses 100\% data) by larger and larger margin (see Rel. Score) as data \% is scaled up.\begin{wraptable}{r}{0.5\linewidth}
\centering
\caption{Scaling performance of \ours.}
 \adjustbox{max width=0.5\columnwidth}{
\begin{tabular}{l c}
\toprule
\textbf{Data Used} & \textbf{Rel. Score (\%)} \\
\midrule
\textbf{Full Finetune} & \\ 
\midrule
100\% & 100 \\
\midrule
\textbf{PROGRESS} & \\ 
\midrule
4.2\% & 89.4 \\
8.3\% & 93.3 \\
16.7\% & 99.0 \\
32.0\% & 100.2 \\
64.5\% & 101.8 \\
\bottomrule
\end{tabular}
}
\label{tab:abl_scaling}
\end{wraptable} \textbf{Reason for better scaling performance}- Our method removes redundancy \& focuses on most informative samples that the model should learn next - naturally shifts attention toward skills that show strong learning potential, instead of spending excessive effort on skills it already performs well on or are too hard to learn at this instant of time.

\subsection{Details for Analysis in main manuscript- How does the benchmark difficulty and data frequency impact performance?}

In this section, we elaborate on details regarding the analysis in Section \ref{sec:analysis}, where we analyze the impact of benchmark difficulty and data frequency.

\begin{algorithm}[t]
\caption{Sample Rarity Estimation via Gaussian Modeling}
\label{alg:rarity}
\begin{algorithmic}[1]
\REQUIRE Training dataset $\mathbb{U} = \{x_1, \ldots, x_N\}$ where $x_i = (I_i, Q_i)$; set of benchmarks $\mathcal{B} = \{B_1, \ldots, B_M\}$ with $M$ different benchmarks.
\STATE \textbf{Feature Extraction:} Extract DINO features from images and BERT features from questions; concatenate to form joint feature vectors, for all training and benchmark samples.
\\ We denote DINO-BERT embedding of $x_i$ as $\phi(x_i)$
\STATE \textbf{Fit Gaussian Models:} Fit multivariate Gaussian $\mathcal{N}(\mu_j, \Sigma_j)$ for each benchmark $B_j$ using its feature vectors
\STATE $n_j \gets 0,\ \forall j \in \{1, \ldots, M\}$ \COMMENT{Initialize match counts for each benchmark}
\FOR{each training sample $x_i \in \mathbb{U}$}
    \STATE $\ell_j = \log \mathcal{N}(\phi(x_i) \mid \mu_j, \Sigma_j), \ \forall j \in \{1, \ldots, M\}$ \COMMENT{Log-likelihoods under each benchmark's Gaussian}
    \STATE $k = \arg\max_j \ell_j$ \COMMENT{Assign to benchmark with highest likelihood}
    \STATE $n_k \gets n_k + 1$ \COMMENT{Increment matched sample count for $B_k$}
\ENDFOR
\FOR{each benchmark $B_j \in \mathcal{B}$}
    \STATE $f_j = n_j / N$ \COMMENT{Compute frequency}
    \STATE $r_j = \log(1 / f_j)$ \COMMENT{Compute rarity}
\ENDFOR
\RETURN $\{r_1, \ldots, r_M\}$ \COMMENT{Rarity scores for all benchmarks}
\end{algorithmic}
\end{algorithm}

\subsubsection{Details for Benchmark Difficulty Analysis}
\label{app:bench_diff}

% {\large \textbf{Details for Benchmark Difficulty Analysis} - }
Here, we provide details regarding the analysis shown in Fig. \ref{fig:difficult_rarity} (a) of main manuscript.
\paragraph{Quantifying Benchmark Difficulty.} Prior work has shown that human intuition about task difficulty may not align with a model’s difficulty as defined in its feature or hypothesis space \citep{sachan2016easy}. Therefore, we use the model’s own performance as a proxy for determining benchmark difficulty. Specifically, we use the performance of full-dataset fine-tuned LLaVA-v1.5-7B (i.e., Row 0 in Table \ref{tab:llava_eval}) as reference to determine difficulty of benchmark—benchmarks with higher performance are considered easier.
We define \textbf{benchmark difficulty} for a given benchmark as $(100 - \text{Performance of full fine-tuned LLaVA-1.5 on Benchmark}) / 100$. This gives us a difficulty measure for each benchmark normalized between $[0,1]$ \footnote{For MME, where the full score is not out of 100, we normalize the score by dividing it by the maximum score (2800), the difficulty is computed as $(1 - \text{MME Score} / 2800)$}.
\paragraph{Quantifying Performance Improvement.}
To isolate the impact of our core contribution—Prioritized Concept Learning (PCL) described in Section~\ref{PCL}—we measure the performance improvement brought solely by our dynamic sample selection strategy. Specifically, we compute the difference in performance between the full \ours framework (Table \ref{tab:ablation_policy}, Row 7) and the warm-up only model trained prior to applying PCL (Table~\ref{tab:ablation_warmup}, Row 1).
This comparison quantifies the gain attributable to dynamically selecting the most informative samples using our PCL strategy during training.

\subsubsection{Details for Data frequency Analysis}
\label{app:data_frequency}
% {\large \textbf{Details for Data frequency Analysis}}
Here, we provide details regarding the analysis shown in Figure \ref{fig:difficult_rarity} (b) of main manuscript.
\paragraph{Sample Rarity Estimation.} Our goal is to identify, for each sample in the training dataset, the benchmark it most closely aligns with in terms of skill distribution.
Each training sample is assigned to exactly one benchmark—whichever it is closest to—based on similarity in distribution over skills. This allows us to estimate the frequency of training samples aligned with each benchmark, enabling us to quantify how commonly each benchmark's skills are represented in the training data.

\paragraph{Assignment Procedure.}  
To quantify how training samples align with various benchmarks, we use a Gaussian modeling approach. Specifically, we first extract visual and textual features using DINO (for
images) and BERT (for questions) and form joint multimodal embeddings as described in Section \ref{sec:MM_cat}—for all samples in training data and each benchmark.

Next, we fit a multivariate Gaussian distribution to each benchmark’s embeddings, capturing its mean and covariance to model the underlying skill distribution. Then, for every training sample, we compute the log-likelihood under each benchmark’s Gaussian model, reflecting how well the sample fits that benchmark's distribution. Each training sample is then assigned to the benchmark with the highest log-likelihood (refer to Algorithm \ref{alg:rarity} for full details).

We compute the frequency of training data samples aligned with each benchmark as the proportion of training samples assigned to it:
\[
\text{frequency} = \frac{\text{\# matched samples}}{\text{total training samples}}.
\]

Finally, we define the rarity as:
% the negative log-frequency:
\[
\text{rarity} = \log(1 / \text{frequency}).
\]

This formulation enables us to assess how frequently the skills associated with each benchmark appear in the training set (see rarity calculation Algorithim \ref{alg:rarity}).

\subsection{Details for Analysis - What Skills Does the Model Prioritize and When?}
\label{app:what_and_when}
  \begin{algorithm}[t]
\caption{Ability Assignment for Clusters}
\label{alg:ability}
\begin{algorithmic}[1]
\REQUIRE Training dataset $\mathbb{U} = \{x_1, \ldots, x_N\}$ where $x_i = (I_i, Q_i)$; Clusters $\mathcal{C} = \{C_1, \dots, C_K\}$; samples from MME benchmark  $\mathcal{B} = \{b_1, \dots, b_M\}$ with ability labels; similarity threshold $\alpha = 0.9$; top‑$k$ nearest neighbors ($k = 5$)

\STATE \textbf{Feature Extraction:} Extract DINO features for images and BERT features for questions; concatenate to form joint feature vectors, for all training and benchmark samples.
\\ We denote DINO-BERT embedding of $x_i$ as $\phi(x_i)$.

\FOR{each cluster $C_k \in \mathcal{C}$}
    \FOR{each sample $x_i \in C_k$}
        % \STATE $\text{sim}(x_i, b_j) = \cos(x_i, b_j), \ \forall b_j \in \mathcal{B}$
        \STATE $\mathcal{N}_i = \{\text{TopK}(\text{sim}(\phi(x_i), \phi(b_j))_{j=1}^M)\}$ $\text{where} \ \text{sim}(\phi(x_i), \phi(b_j)) = \cos(\phi(x_i), \phi(b_j))$ 
        \STATE $\mathcal{N}_i^\prime = \{b_j \in \mathcal{N}_i \mid \text{sim}(\phi(x_i), \phi(b_j)) \geq \alpha \cdot \max_j \text{sim}(\phi(x_i), \phi(b_j))\}$
        \STATE $\mathcal{A}_i = \{\text{ability}(b_j) \mid b_j \in \mathcal{N}_i^\prime\}$
    \ENDFOR
    \STATE $\mathcal{A}_k = \bigcup_{x_i \in C_k} \mathcal{A}_i$ \COMMENT{Aggregate ability labels from all samples in $C_k$}
    \STATE $\text{Ability}(C_k) = \text{Mode}(\mathcal{A}_k)$ \COMMENT{Assign the most frequent ability via majority vote}

\ENDFOR

\RETURN $\{ \text{Ability}(C_1), \dots, \text{Ability}(C_K) \}$
\end{algorithmic}
\end{algorithm}

In this section, we elaborate on the analysis from Section~\ref{sec:analysis} (specifically Fig. \ref{fig:ability_learning} in main manuscript), where we investigate which skills the model prioritizes and when during training.

Our goal is to identify the specific ability each skill-cluster—obtained through our concept categorization module described in Section~\ref{sec:MM_cat})—represents and track both the number of selected samples and the performance of that skill over time. To do this, we assign each skill cluster in our framework to one of the standardized ability categories defined by the MME benchmark (i.e \textit{count}, \textit{position}, \textit{OCR} etc) which offer interpretable and fine-grained labels covering both perception and cognitive tasks. To determine the dominant ability for each skill-cluster , we use a similarity-based assignment procedure (see Algorithm \ref{alg:ability} for details).

We first extract visual and textual features using DINO (for
images) and BERT (for questions) and form joint multimodal embeddings as described in Section \ref{sec:MM_cat}—for all samples in training data and MME benchmark dataset.

For each training sample in a given skill-cluster generated by our concept categorization module, we compute its cosine similarity with all samples in the MME benchmark. We identify its top-$K$ nearest neighbors in MME benchmark and retain only those with similarity above a 90\% threshold, ensuring that we capture the most aligned MME samples for each training example. MME abilities associated with these filtered neighbors are aggregated for samples in the cluster, and majority voting is applied to assign the most frequent ability to the entire cluster. This process offers a principled way to characterize each skill-cluster’s dominant visual-linguistic ability, ensuring robustness through both similarity filtering and voting (refer to Algorithim \ref{alg:ability} for more details).

% \vspace{-2mm}
\section{Analysis}
% \vspace{-2mm}
\subsection{Word Cloud Visualization of Skill Clusters}
\label{app:word_cloud}
To qualitatively assess the semantic coherence and purity of discovered skill clusters obtained through our concept categorization module (Section~\ref{sec:MM_cat}), we generate word clouds by aggregating all questions from all samples assigned to a given cluster. For each cluster, we concatenate all the corresponding questions into a single string and visualize the most frequent words using wordclouds. Note that we remove standard stopwords while plotting the wordclouds.

\begin{figure}[t]
    \centering
    \begin{subfigure}[b]{0.32\linewidth}
        \includegraphics[width=\linewidth]{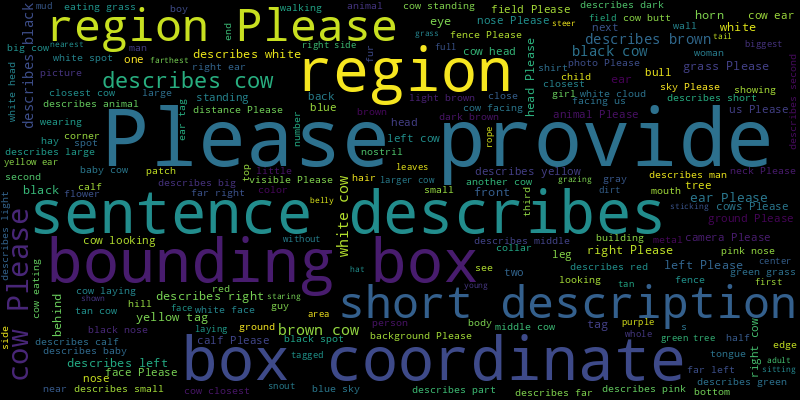}
        \caption{ }
    \end{subfigure}
    \begin{subfigure}[b]{0.32\linewidth}
        \includegraphics[width=\linewidth]{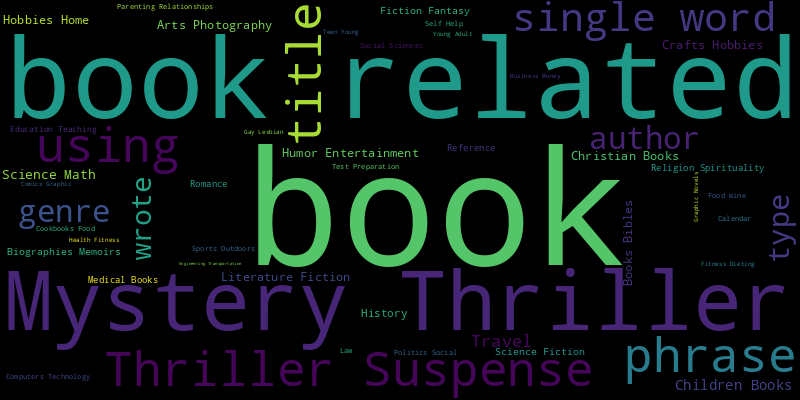}
        \caption{ }
    \end{subfigure}
    \begin{subfigure}[b]{0.32\linewidth}
        \includegraphics[width=\linewidth]{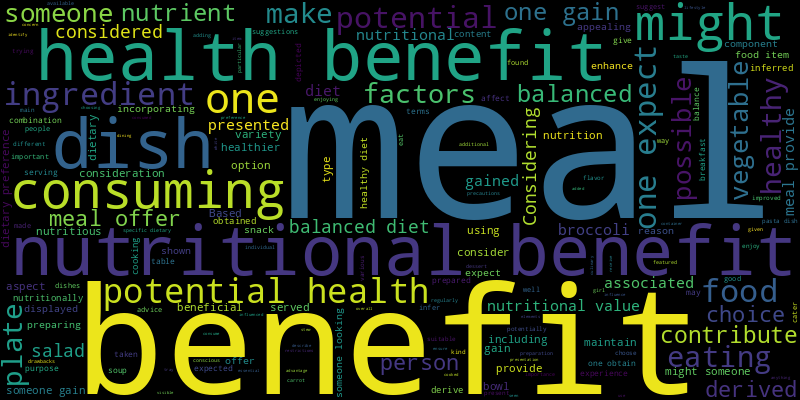}
        \caption{ }
    \end{subfigure}
    \\
    \begin{subfigure}[b]{0.32\linewidth}
        \includegraphics[width=\linewidth]{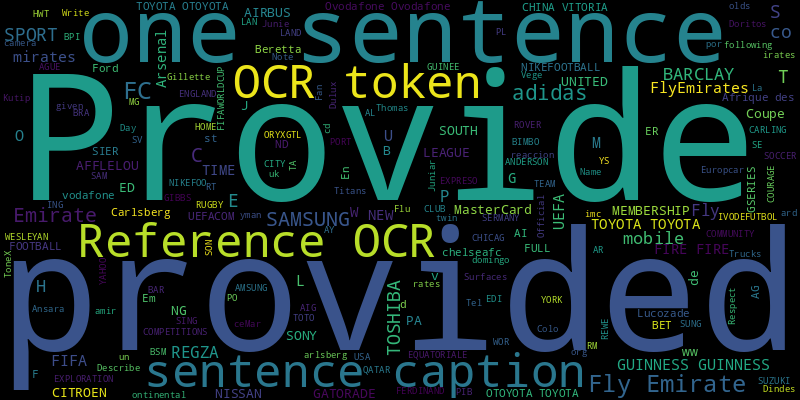}
        \caption{ }
    \end{subfigure}
    \begin{subfigure}[b]{0.32\linewidth}
        \includegraphics[width=\linewidth]{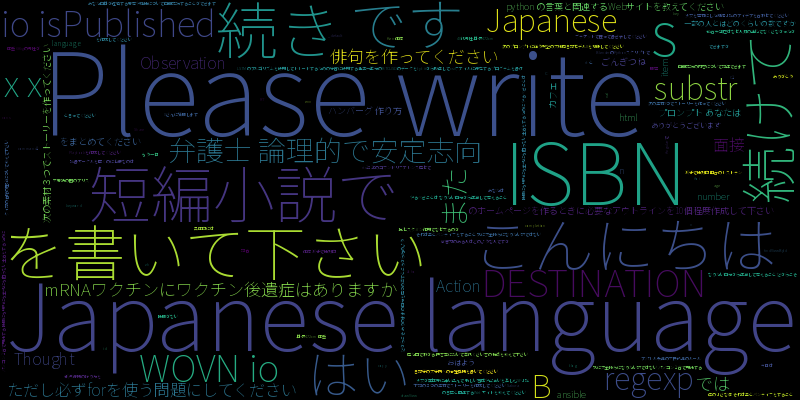}
        \caption{ }
    \end{subfigure}
    \begin{subfigure}[b]{0.32\linewidth}
        \includegraphics[width=\linewidth]{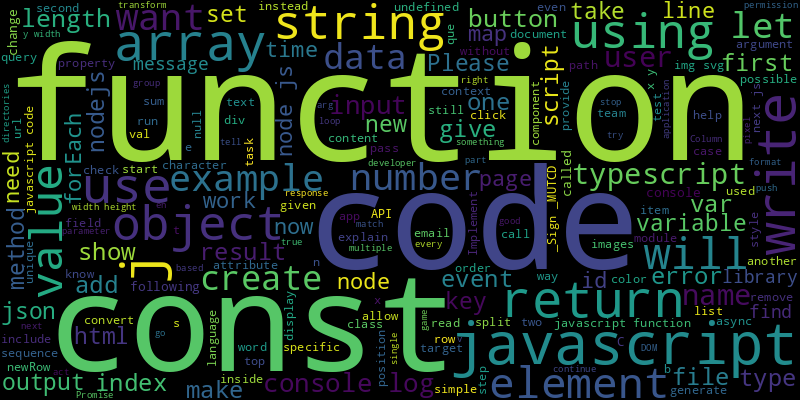}
        \caption{ }
    \end{subfigure}
    \caption{{\textbf{Word Cloud Visualization of Skill Clusters.}  
Each subfigure shows the word cloud generated by concatenating all questions within a single skill-cluster discovered by our unsupervised concept categorization module. The clusters exhibit clear semantic themes: (a) object localization and region descriptions, (b) book metadata and genres, (c) food and nutritional benefits, (d) OCR, (e) multilingual Japanese text and language prompts, and (f) programming and function-related instructions. These word clouds highlight the semantic coherence and fine-grained granularity of the automatically discovered clusters, validating their utility for skill-level progress tracking.}
}
    \label{fig:word_cloud}
    \vspace{-3mm}
\end{figure}

Figure~\ref{fig:word_cloud} shows representative word clouds for six clusters. Each cluster exhibits a distinct semantic theme, validating the purity and fine-grained granularity of the automatically discovered clusters and demonstrating the effectiveness of our multimodal concept categorization. For example, cluster (a) object localization and region descriptions, (b) book metadata and genres, (c) pertains to food and nutritional benefits, (d) corresponds to OCR and reference tokens, (e) involves multilingual Japanese text and language prompts, and (f) highlights programming and function-related tasks.

These visualizations demonstrate that our clustering method forms fine-grained, interpretable concept groupings while being fully unsupervised (see Section \ref{sec:MM_cat})—essential for skill-level tracking and prioritized learning in PROGRESS.

% \vspace{-2mm}
\subsection{Skill-level Diversity in Selected Sample Distribution}
\begin{figure*}[b]
    \centering
    \begin{minipage}{\textwidth}
        \centering
        \includegraphics[width=0.7\linewidth]{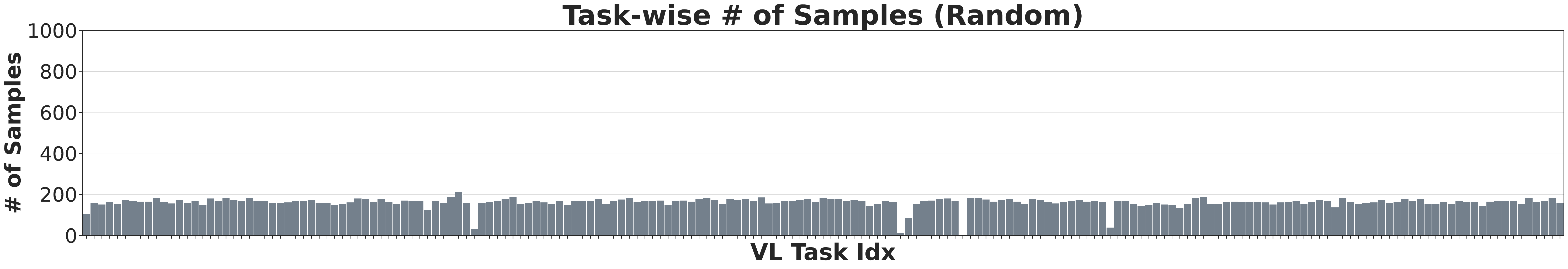}
    \end{minipage}
    \par
    \vspace{0.05in}
    \begin{minipage}{\textwidth}
        \centering
        \includegraphics[width=0.7\linewidth]{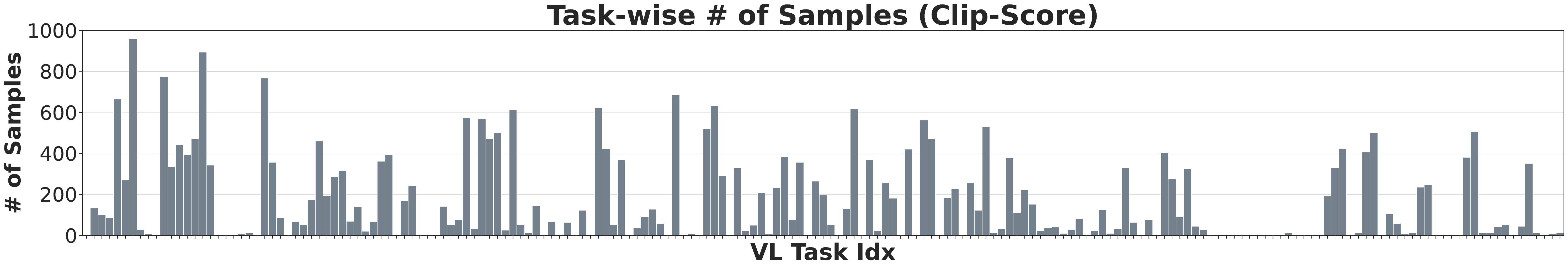}
    \end{minipage}
    \par
    \vspace{0.05in}
    \begin{minipage}{\textwidth}
        \centering
        \includegraphics[width=0.7\linewidth]{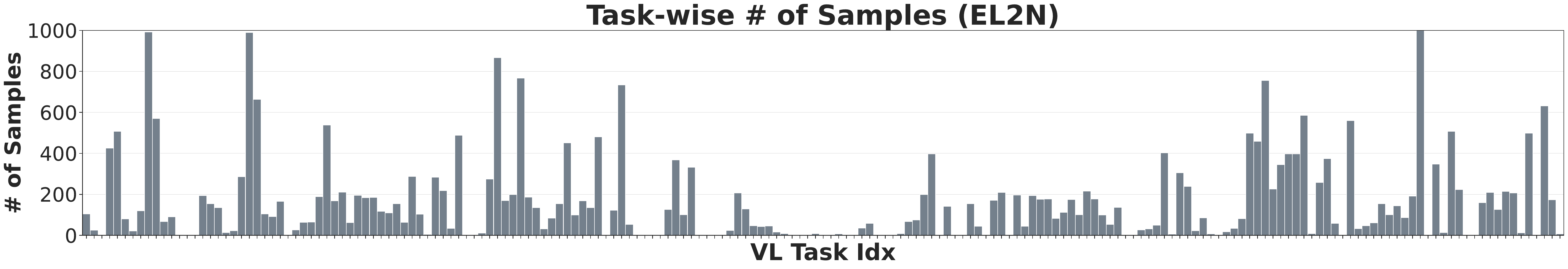}
    \end{minipage}
    \par
    \vspace{0.05in}
    \begin{minipage}{\textwidth}
        \centering
        \includegraphics[width=0.7\linewidth]{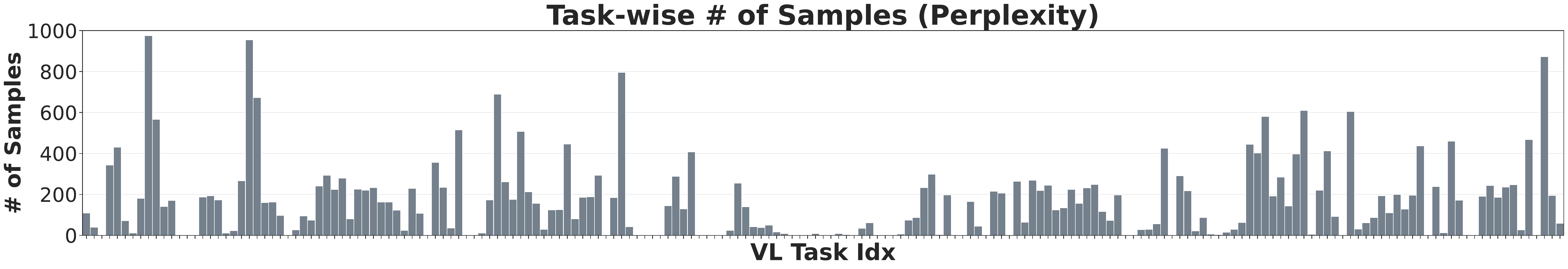}
    \end{minipage}
    \par
    \vspace{0.05in}
    \begin{minipage}{\textwidth}
        \centering
        \includegraphics[width=0.7\linewidth]{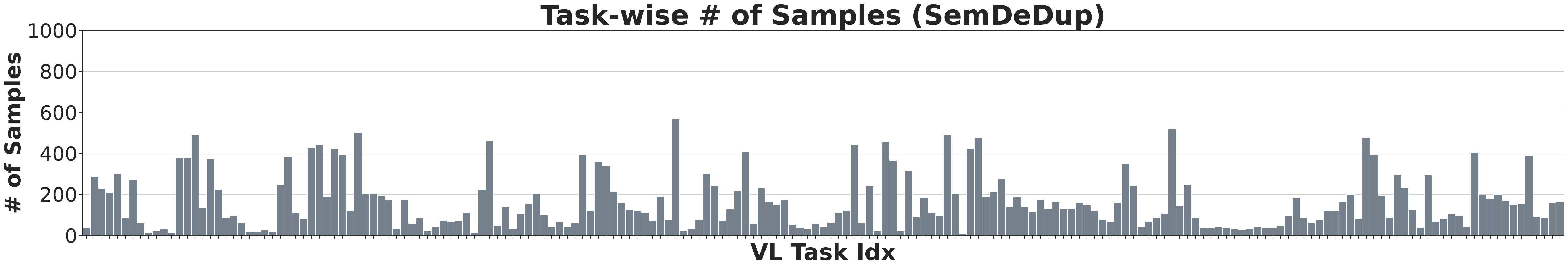}
    \end{minipage}
    \par
    \vspace{0.05in}
    \begin{minipage}{\textwidth}
        \centering
        \includegraphics[width=0.7\linewidth]{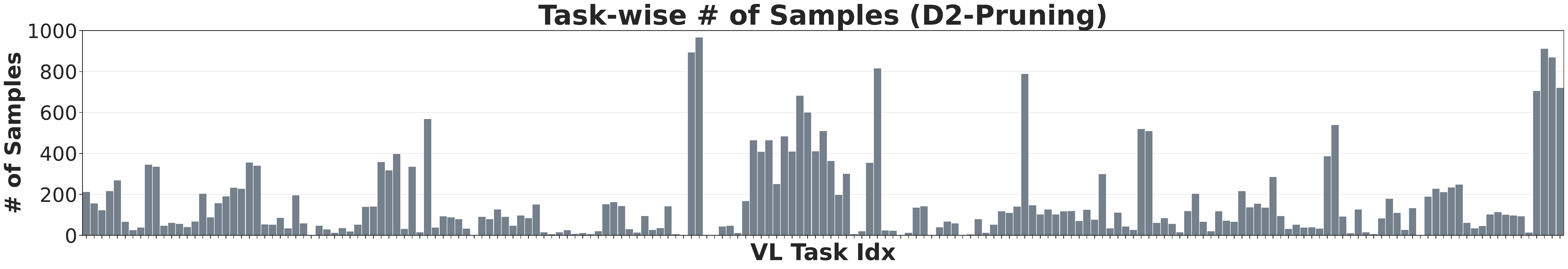}
    \end{minipage}
    \par
    \vspace{0.05in}
    \begin{minipage}{\textwidth}
        \centering
        \includegraphics[width=0.7\linewidth]{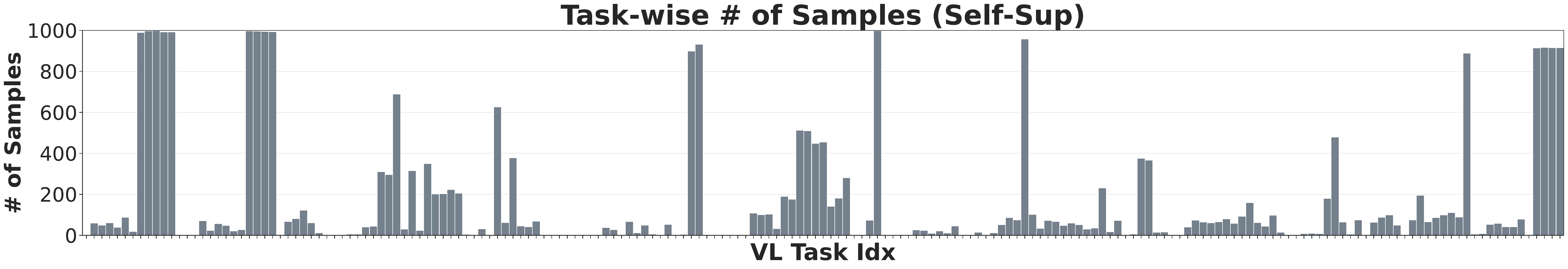}
    \end{minipage}
    \par
    \vspace{0.05in}
    \begin{minipage}{\textwidth}
        \centering
        \includegraphics[width=0.7\linewidth]{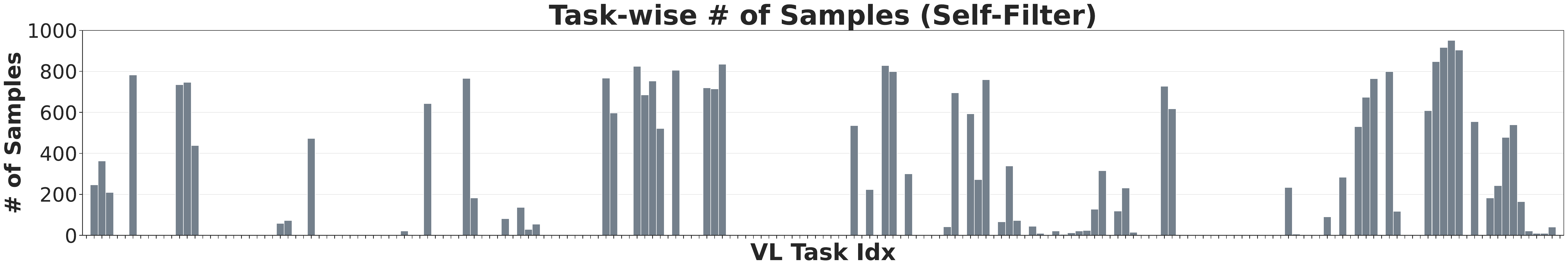}
    \end{minipage}
    \par
    \vspace{0.05in}
    \begin{minipage}{\textwidth}
        \centering
        \includegraphics[width=0.7\linewidth]{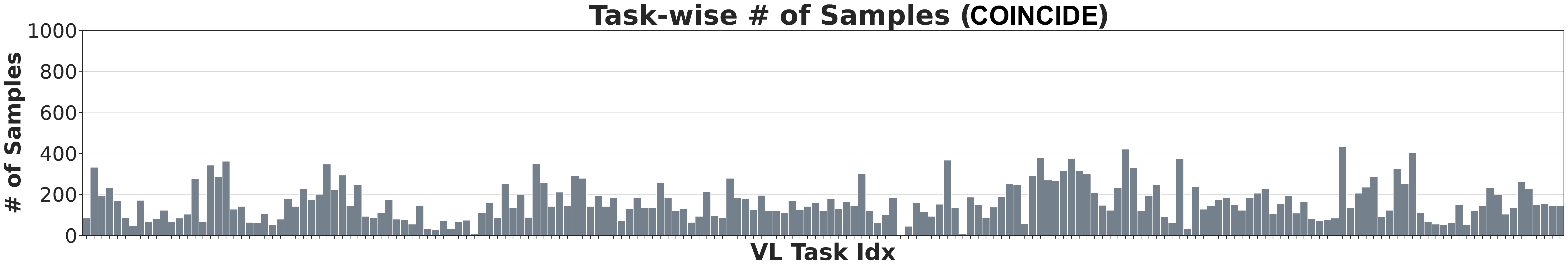}
    \end{minipage}
    \par
    \vspace{0.05in}
    \begin{minipage}{\textwidth}
        \centering
        \includegraphics[width=0.7\linewidth]{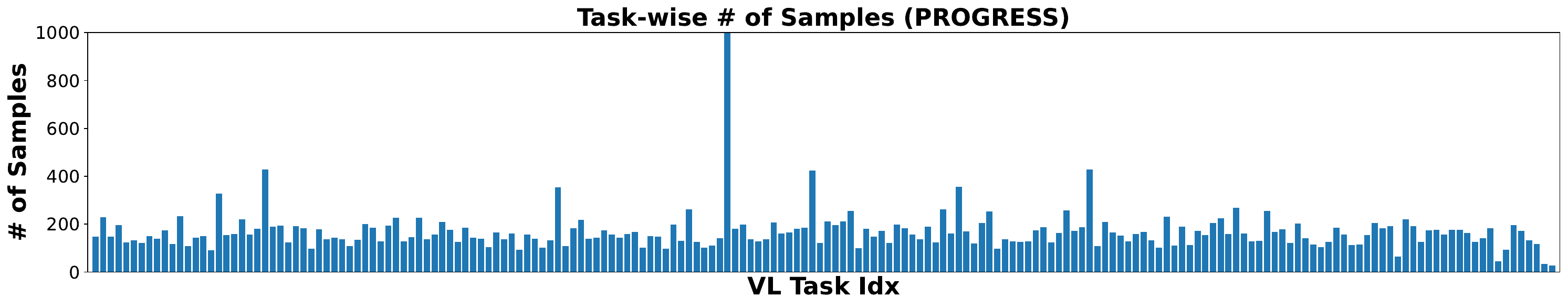}
    \end{minipage}
    \par
    \vspace{0.05in}
    \captionsetup{justification=justified}
    \caption{\textbf{Task-wise Distribution of Selected Samples.} Number of samples selected (y-axis) from each Vision-Flan-191K task (x-axis) across different methods. While baselines tend to concentrate heavily on a few high-scoring tasks, PROGRESS achieves a more balanced sampling pattern across the task spectrum—highlighting its ability to maintain skill diversity.
\label{fig:task_distribution}}

    \label{fig:supple_vision_flan_diversity}
    % \vspace{-0.075in}
\end{figure*}

% \vspace{-2mm}
To better understand the selection behavior across methods, we follow protocol in previous work \citep{coincide} and analyze the number of selected samples from each task in the Vision-Flan-191K dataset. Figure~\ref{fig:task_distribution} shows the task-wise sample distribution for \ours and several baseline approaches.

We observe that methods relying on single static scoring functions—such as CLIP-Score, EL2N, Perplexity, and Self-Sup—tend to exhibit strong sampling bias, disproportionately selecting from a small subset of tasks while neglecting others. This narrow focus often overlooks important data modes, leading to poor generalization—a limitation also noted in prior work~\citep{coincide}.

In contrast, \ours maintains a more balanced and diverse sampling profile across tasks, ensuring that a broader range of skills and task types are represented during training. This diversity stems from our skill-driven selection strategy, which tracks learning progress across clusters and samples proportionally using a temperature-controlled distribution.

Overall, by avoiding the pitfalls of static scoring and overfitting to specific high-scoring skills or frequent tasks, our method instead  promotes broader and more effective skill acquisition.

\section{Limitations}
\label{sec:limitation}

While \ours effectively orders and prioritizes more informative skills, it randomly samples within each selected skill cluster without ranking samples by usefulness. Additionally, the accuracy-based variant incurs extra inference time to compute skill-level progress (see Appendix \ref{app:exp}), though our loss-based variant avoids this issue. However, overall, \ours outperforms prior approaches while requiring no additional reference VLM and significantly less supervised data.

\section{LLM Usage Disclosure}

The LLM was primarily used for language refinement rather than any content generation—all experimental designs, results, analyses, and scientific contributions are original work by the authors. The LLM assistance was limited to editorial improvements such as fixing grammatical errors, suggesting clearer phrasing for complex technical concepts, and ensuring consistency in terminology throughout the manuscript. No experimental results, mathematical derivations, or scientific claims were generated by the LLM. All factual statements, citations, and technical content were independently verified by the authors.

\newpage

\newpage

\end{document}